\documentclass{ieeetj}

\usepackage{algpseudocode}
\usepackage[linesnumbered, ruled, vlined]{algorithm2e}
\usepackage{array}
\usepackage[caption=false,font=normalsize,labelfont=sf,textfont=sf]{subfig}
\usepackage[colorlinks,urlcolor=blue,linkcolor=blue,citecolor=blue]{hyperref}
\usepackage{textcomp}
\usepackage{stfloats}
\usepackage{url}
\usepackage{multirow}
\usepackage{verbatim}
\usepackage{booktabs}
\usepackage{graphicx}
\usepackage[noadjust]{cite}

\usepackage{amsmath,amssymb,amsfonts}
\usepackage{graphicx,color}
\usepackage{textcomp}
\usepackage{xcolor}

\usepackage{xr-hyper}
\usepackage[moderate,tracking=normal,paragraphs=normal]{savetrees} 

\def\BibTeX{{\rm B\kern-.05em{\sc i\kern-.025em b}\kern-.08em
    T\kern-.1667em\lower.7ex\hbox{E}\kern-.125emX}}
\AtBeginDocument{\definecolor{tmlcncolor}{cmyk}{0.93,0.59,0.15,0.02}\definecolor{NavyBlue}{RGB}{0,86,125}}

\usepackage{xcolor,soul,framed} 
\colorlet{shadecolor}{yellow}

\def\OJlogo{\vspace{-4pt}$<$Society logo(s) and publication title will appear here.$>$}
\def\seclogo{\vspace{10pt}$<$Society logo(s) and publication title will appear here.$>$}

\def\authorrefmark#1{\ensuremath{^{\textbf{#1}}}}

\begin{document}
\receiveddate{XX Month, XXXX}
\reviseddate{XX Month, XXXX}
\accepteddate{XX Month, XXXX}
\publisheddate{XX Month, XXXX}
\currentdate{XX Month, XXXX}
\doiinfo{XXXX.2022.1234567}

\markboth{Human-in-the-Loop Feature Selection Using Interpretable Kolmogorov-Arnold Network-based Double Deep Q-Network}{Jahin {et al.}}

\title{Human-in-the-Loop Feature Selection Using Interpretable Kolmogorov-Arnold Network-based Double Deep Q-Network}

\author{Md Abrar Jahin\authorrefmark{1},~\IEEEmembership{Graduate Student Member,~IEEE,}
          M.~F.~Mridha\authorrefmark{2},~\IEEEmembership{Senior Member,~IEEE,}
          Nilanjan~Dey\authorrefmark{3},~\IEEEmembership{Senior Member,~IEEE}
          and Md.~Jakir~Hossen\authorrefmark{4},~\IEEEmembership{Senior Member,~IEEE}
}
\affil{Thomas Lord Department of Computer Science, Viterbi School of Engineering, University of Southern California, Los Angeles, CA 90089, USA (e-mail: jahin@usc.edu)}
\affil{Department of Computer Science, American International University-Bangladesh, Dhaka 1229, Bangladesh (e-mail: firoz.mridha@aiub.edu)}
\affil{Department of Computer Science \& Engineering, Techno International New Town, New Town, Kolkata 700156, India (e-mail: nilanjan.dey@tint.edu.in)}
\affil{Department of Robotics and Automation, Multimedia University, 75450 Bukit Beruang, Melaka, Malaysia (e-mail: jakir.hossen@mmu.edu.my)}

\corresp{Corresponding author(s): Md. Jakir Hossen(e-mail: jakir.hossen@mmu.edu.my), Md Abrar Jahin (e-mail: jahin@usc.edu), and M. F. Mridha (e-mail: firoz.mridha@aiub.edu).}

\begin{abstract}
Feature selection is critical for improving the performance and interpretability of machine learning models, particularly in high-dimensional spaces where complex feature interactions can reduce accuracy and increase computational demands. Existing approaches often rely on static feature subsets or manual intervention, limiting adaptability and scalability. However, dynamic, per-instance feature selection methods and model-specific interpretability in reinforcement learning remain underexplored. This study proposes a human-in-the-loop (HITL) feature selection framework integrated into a Double Deep Q-Network (DDQN) using a Kolmogorov-Arnold Network (KAN). Our novel approach leverages simulated human feedback and stochastic distribution-based sampling, specifically Beta, to iteratively refine feature subsets per data instance, improving flexibility in feature selection. The KAN-DDQN achieved notable test accuracies of 93\% on MNIST and 83\% on FashionMNIST, outperforming conventional MLP-DDQN models by up to 9\%. The KAN-based model provided high interpretability via symbolic representation while using 4 times fewer neurons in the hidden layer than MLPs did. Comparatively, the models without feature selection achieved test accuracies of only 58\% on MNIST and 64\% on FashionMNIST, highlighting significant gains with our framework. We further validate scalability on CIFAR-10 and CIFAR-100, achieving up to 30\% relative macro F1 improvement on MNIST and 5\% on CIFAR-10, while reducing calibration error by 25\%. Complexity analysis confirms real-time feasibility with latency below 1\,ms and parameter counts under 0.02M. Pruning and visualization further enhanced model transparency by elucidating decision pathways. These findings present a scalable, interpretable solution for feature selection that is suitable for applications requiring real-time, adaptive decision-making with minimal human oversight. The code is available at \url{https://github.com/Abrar2652/HITL-FS}.
\end{abstract}


\maketitle

\section{Introduction}
\IEEEPARstart{F}{eature} selection is essential for building efficient, accurate, and robust machine learning models~\cite{guyon_introduction_2003,chandrashekar_survey_2014}. While models ideally should automatically identify the most predictive features, a high-dimensional input space can significantly hinder performance, often requiring large volumes of data to effectively learn the complex relationships between features. This phenomenon, known as the ``curse of dimensionality," increases computation time and resource use. Consequently, selecting a smaller subset of relevant features improves performance and makes the model more cost-effective.

One common solution is to incorporate expert knowledge to determine the most useful features; however, this process is costly, time-consuming, and highly manual. Additionally, experts with deep domain knowledge are often not involved in the actual design and development of the model. Automatic feature selection methods offer an alternative, ranking features by their relevance or importance; however, both manual and automatic approaches typically yield a single subset of features for the entire dataset, which may not capture variability across individual observations. When training data are sparse relative to the feature space, using a single subset can limit the model's ability to generalize effectively across all instances~\cite{correia_human---loop_2019}.

To address the challenge of per-example feature selection, we propose a reinforcement learning (RL) framework that leverages simulated feedback to replicate human feature selection. Our approach employs a Double Deep Q-Network (DDQN) setup with a Kolmogorov-Arnold Network (KAN)~\cite{kich_kan, guo_kan} as both the Q-network and the target network. This model-specific, interpretable KAN-based RL structure aims to refine feature selection on a per-example basis iteratively. In this setup, simulated feedback, rather than direct human input, is a proxy for expert annotation~\cite{cano_method_2011,daee_knowledge_2017}. The feedback signal highlights the most relevant features for each data example, enabling the model to prioritize these features during training. Unlike prior methods~\cite{correia_human---loop_2019}, our model explores distributions beyond Bernoulli and organizes convolutional and pooling layers to ensure feature maps match the input and simulated feedback shapes. RL then optimizes a policy to select a unique subset of features per observation. By minimizing the classifier’s prediction loss and the discrepancy between simulated feedback and the model-selected features, this policy yields feature subsets that improve the interpretability and performance of the final predictions. Since predictions are based only on the selected feature subsets, this method offers interpretable, case-specific insights into the model’s output. Using simulated feedback further enables the model to reflect causal structures likely to be relevant in practical applications. We validate our methodology through rigorous experimentation on benchmark datasets, showing the efficiency of our approach in improving model accuracy while maintaining computational feasibility. We aim to establish best practices for integrating human feedback into the feature selection process by investigating the influence of various hyperparameters, stochastic distributions, and the absence of feature selection.

In many deployed classification settings, such as safety triage in vision, early exits in mobile perception, and clinician-in-the-loop screening, models must meet tight memory/latency budgets while preserving case-wise interpretability. Static, dataset-level feature subsets often underperform when intra-class variability is high; conversely, per-instance selection can reduce redundant computation and expose the causal drivers of individual decisions. Our goal is to operationalize \emph{instance-wise} feature selection under realistic constraints: (i) a compact head suitable for resource-limited environments, (ii) a stochastic, differentiable gating mechanism that trades sparsity for accuracy, and (iii) a reinforcement-learning controller whose policy remains interpretable. This motivates our HITL-DDQN framework, which features a KAN or MLP head and a stochastic gate aligned with simulated feedback, thereby delivering actionable sparsity–accuracy operating points and model-specific interpretability.

The key components of our contribution include the following:
\begin{enumerate}
\item We introduce a novel approach incorporating simulated feedback via Gaussian heatmaps and stochastic, distribution-based sampling to refine feature subsets on a per-example basis, thereby enhancing model interpretability and performance.
\item By incorporating KAN into the DDQN architecture for both the Q-network and target network, we achieve significantly better performance than traditional MLP-based DDQN across all test cases. This approach uses a hidden layer with four times fewer neurons than MLP while offering model-specific interpretability.
\item Our research presents a simulated feedback mechanism that generates feature relevance feedback without the need for human annotators, facilitating a scalable training process that reflects the causal relationships typically identified by human experts.
\end{enumerate}

The following sections provide an overview of our work: Section \ref{sec:2} reviews the relevant background; Section \ref{sec:3} outlines the complete methodology of our proposed framework; Section \ref{sec:4} details the experimental design; Section \ref{sec:5} discusses the results and their interpretations; and finally, Section \ref{sec:6} concludes the research while outlining potential future directions.

\section{Background}\label{sec:2}
\subsection{Human-in-the-Loop Feature Selection Using RL}
Feature selection is critical in developing machine learning models but is often executed through data-driven methods that overlook insights from human designers~\cite{chandrashekar_survey_2014,kumar_applications_2024}. We introduce a HITL~\cite{wu_survey_2022} framework that integrates simulated feedback to identify the most relevant variables for specific tasks, which can be modeled using DDQN-based~\cite{hasselt_deep_2016} RL to facilitate per-example feature selection~\cite{raghavan_active_2006}, enabling the model to minimize its loss function while emphasizing significant variables from a simulated human perspective. A major gap in RL is the limited model-specific interpretability, as most models operate as black-box systems, making it challenging to understand their decision-making processes~\cite{verma_programmatically_2018}. Our KAN agent enhances the interpretability of DDQN-based RL by providing symbolic representations of learned policies. Our methodology employs variable elimination techniques, focusing on selecting subsets of features rather than using embedded methods. This approach optimizes feature selection and learning processes concurrently via gradient descent, distinguishing itself by selecting different subsets for each observation. This enhances interpretability, as the chosen variables represent the ``causes" driving the model's predictions. While previous work \cite{guyon_introduction_2003,wu_survey_2022} has explored per-example feature selection via traditional filter methods based on mutual information, our framework goes beyond incorporating simulated human feedback for a more dynamic approach. Traditional feature selection techniques include filter methods, which select top-ranked features based on criteria like mutual information; wrapper methods, which evaluate subsets of features by retraining models for each subset; and embedded methods, which attempt to select features during the model training process. Unlike these approaches, which can be sequential and computationally costly, our model generates candidate subsets in a single step, guided by simulated feedback, thereby avoiding arbitrary stopping criteria and allowing for real-time, per-example selection in complex datasets. Our work draws inspiration from the probabilistic knowledge elicitation \cite{cano_method_2011,wu_finite-horizon_2024} focused on querying users for global feature relevance. However, our model goes beyond the focus on Bernoulli distributions \cite{correia_human---loop_2019} by exploring various distributions, including beta, Gaussian, and Gumbel-Softmax, enhancing the flexibility and effectiveness of feature selection. This breadth allows us to capture a wider range of relationships within the data compared with previous approaches.

\subsection{Kolmogorov-Arnold Networks (KANs)}\label{back:kan}
KANs~\cite{liu_kan_2024} utilize the Kolmogorov-Arnold representation theorem, which states that any multivariate continuous function \( f: [0, 1]^n \to \mathbb{R} \) can be expressed as a finite composition of univariate functions and additions. Mathematically, this is formulated as follows:
\begin{equation}
f(x) = \sum_{q=1}^{2^{n+1}} \Phi_q \left( \sum_{p=1}^{n} \phi_{q,p}(x_p) \right)
\end{equation}
where \( \phi_{q,p}: [0, 1] \to \mathbb{R} \) and \( \Phi_q: \mathbb{R} \to \mathbb{R} \) are continuous functions. In KANs, the traditional weight parameters are substituted with learnable one-dimensional B-spline functions \( \phi \). The output of a KAN layer with \( n_{in} \) inputs and \( n_{out} \) outputs is given by:
\begin{equation}
x_{l+1,j} = \sum_{i=1}^{n_l} \phi_{l,j,i}(x_{l,i})
\end{equation}
where \( \phi_{l,j,i} \) connects the \( i \)-th neuron in layer \( l \) to the \( j \)-th neuron in layer \( l+1 \). The backpropagation algorithm computes gradients with respect to the spline coefficients \( c_i \) to minimize the loss \( L \) via gradient descent:
\begin{equation}
\frac{\partial L}{\partial c_i} = \sum_{j=1}^{n_{out}} \frac{\partial L}{\partial x_{l+1,j}} \frac{\partial x_{l+1,j}}{\partial c_i}
\end{equation}
where \( \frac{\partial x_{l+1,j}}{\partial c_i} \) reflects the derivative of the spline function concerning its coefficients. KANs define spline functions \( \phi \) over a discretized grid, which specifies where the functions are evaluated, thus influencing the approximation resolution. The order parameter indicates the degree of the B-splines: an order of 1 denotes linear splines, while higher orders yield more complex shapes. The grid and order parameters determine KANs' capacity to model intricate functional relationships; finer grids with higher-order splines allow for precise approximations but require more computational resources.

The primary difference between KAN and MLP is that KANs implement learnable activation functions along edges, whereas MLPs implement fixed activation functions at nodes. KANs represent weights as splines, improving their ability to approximate complex functions with fewer parameters. KANs and MLPs can be extended to multiple layers, supporting deep architectures. We introduce KANs here to frame their conceptual differences from MLPs; the practical instantiation inside our DDQN head (width, grid, regularization) is given later in Sec. {\ref{sec:3}}-{\ref{method:kan}} to avoid duplication and to keep implementation choices distinct from background theory.

\begin{figure*}[!ht]
\centering
\includegraphics[width=\linewidth]{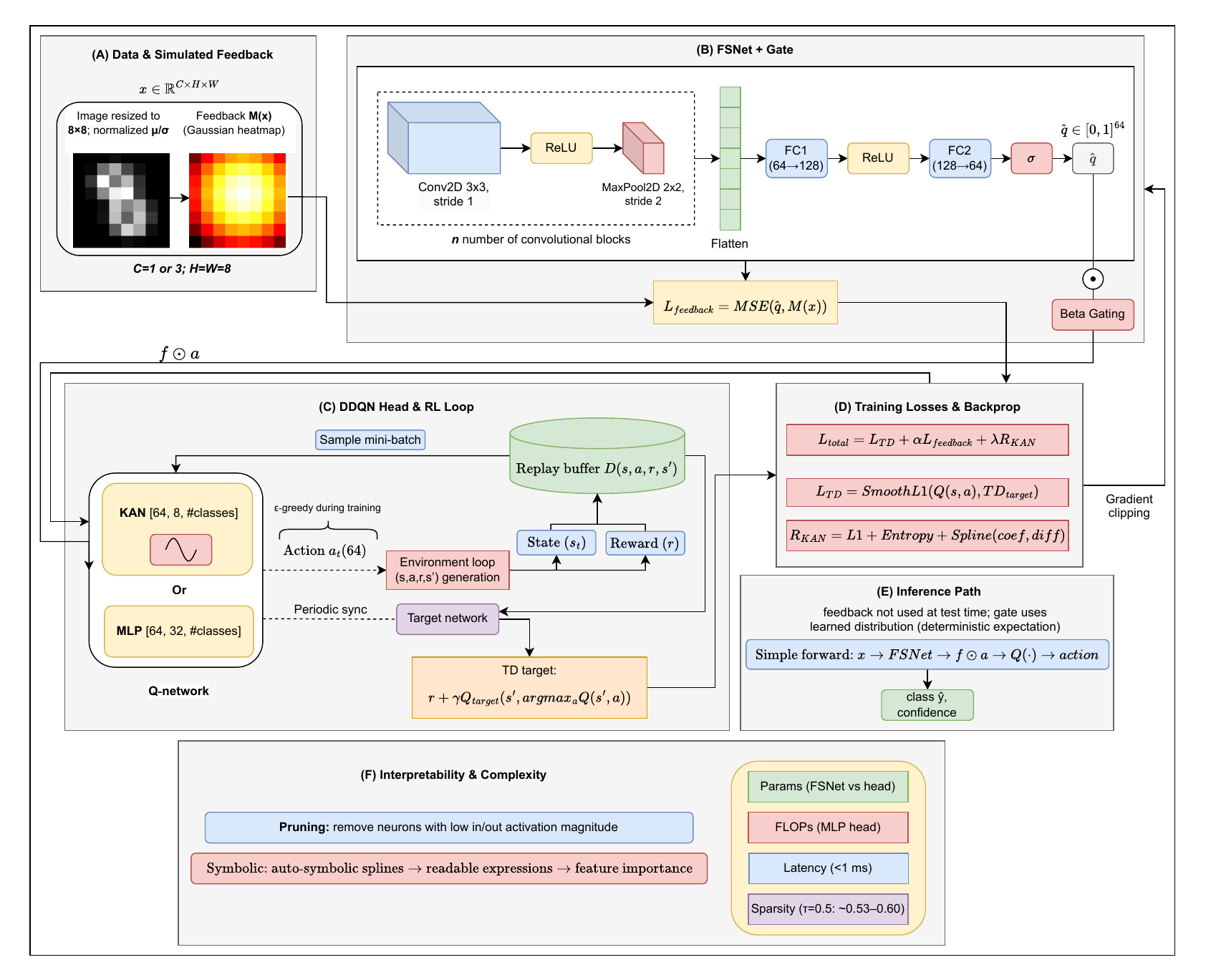}
\caption{System overview. The feature-selection head (FSNet) maps an input image to per-feature probabilities via a differentiable stochastic gate aligned to simulated feedback. The selected features are fed to a DDQN head (KAN or MLP). Replay-based training updates the Q-network, with periodic target sync. The gate induces instance-wise sparsity and exposes feature-level rationale, while the KAN head provides model-specific interpretability.}
\label{fig:system}
\end{figure*}

\section{Methodology}\label{sec:3}

\begin{algorithm}[!ht]
\footnotesize
\label{algo:1}
\caption{FSNet Model Initialization and Forward Pass}
\KwIn{Input shape $input\_shape$, Number of filters $num\_filters$, Number of convolutional layers $num\_conv\_layers$, Hidden layer dimension $hidden\_dim$, Distribution $distribution$, Temperature parameter $\tau$}
\KwResult{Processed feature map, feedback cost, feature probabilities}
\BlankLine

\SetKwFunction{FMain}{FSNet}
\SetKwProg{Fn}{Function}{:}{}
\Fn{\FMain{$input\_shape$, $num\_filters$, $num\_conv\_layers$, $hidden\_dim$, $distribution$, $\tau$}}{
    Initialize convolutional layer parameters: $channels \gets $\ [];
    \For{$i \gets 1$ \KwTo $num\_conv\_layers$}{
        $num\_filters \gets num\_filters \times 2$\;
        Append $num\_filters$ to $channels$\;
    }
    Build feature extractor with $num\_conv\_layers$ Conv2D, ReLU, and MaxPool layers\;
    Initialize fully connected layer $fc1$ with input size $n\_features$ and hidden size $hidden\_dim$\;
    Initialize output layer $fc2$ with input size $hidden\_dim$ and output size $n\_features$\;
    Initialize activation functions: $relu$, $sigmoid$, and $softmax$\;
}
\BlankLine

\SetKwFunction{FForward}{forward}
\SetKwProg{Fn}{Function}{:}{\KwRet}
\Fn{\FForward{$x$, $feedback$, $epoch$}}{
    Reshape $x$ to match $input\_shape$\;
    $features \gets$ Pass $x$ through feature extractor\;
    $x \gets relu(fc1(features))$\;
    Compute $logits \gets relu(fc2(x))$\;
    Compute feature selection probabilities $probs \gets sigmoid(logits)$\;
    
    \If{$feedback$ is None}{
        $feedback \gets probs$\;
    }
    
    Sample $sample\_probs$ from $distribution$ using $logits$ as parameters\;
    Adjust temperature parameter if Gumbel-Softmax is used: $tau \gets \tau \times \tau\_decay^{epoch}$\;

    Calculate feedback cost $feedback\_cost \gets MSE\_Loss(probs, feedback)$\;
    $selected\_feature \gets features \times sample\_probs$\;

    \KwRet{$selected\_feature$, $feedback\_cost$, $probs$}\;
}
\end{algorithm}

\subsection{Human-in-the-Loop Feature Selection Process}
The feature selection network (FSNet) (see Algorithm \ref{algo:1} and Fig. \ref{fig:system}) is designed to determine the relevance of each input feature for a given task~\cite{raghavan_active_2006}. The architecture combines convolutional layers for feature extraction with various probabilistic distributions to model and sample feature importance. Here, we detail each model component, focusing on architectural elements, distribution-based sampling, and feedback alignment.

\subsubsection{Convolutional Feature Extraction Layers}
The first stage of FSNet involves convolutional layers that transform the input tensor \( x \) into feature representations that capture important patterns across spatial dimensions. Suppose \( x \in \mathbb{R}^{d \times h \times w} \) represents an input with \( d \) channels, height \( h \), and width \( w \). The convolutional feature extractor uses \( L \) layers of convolution, ReLU activation, and max-pooling to refine the spatial features of the input progressively. For layer \( i \), with filter count \( f_i \), the transformation can be expressed as:

\begin{equation}
F^{(i)} = \text{MaxPool2d}(\text{ReLU}(\text{Conv2d}(F^{(i-1)})))
\end{equation}
where \( F^{(i)} \in \mathbb{R}^{f_i \times \frac{h}{2^i} \times \frac{w}{2^i}} \) and \( F^{(0)} = x \). Each convolutional layer uses a $3\times3$ kernel and padding of 1 to maintain spatial dimensions, while max-pooling layers with a kernel size of 2 reduce the height and width by half at each step.

\subsubsection{Fully Connected Layers for Feature Probability Mapping}
After convolutional layers, the resulting feature map is flattened and processed by fully connected layers to output feature relevance scores (logits). Let \( F \in \mathbb{R}^{n} \) represent the flattened feature vector, where \( n \) is the number of features output by the convolutional layers. This vector is then passed through a sequence of fully connected (FC) layers, ReLU activation, and dropout:
\begin{equation}
h = \text{ReLU}(W_{\text{fc1}} \cdot F + b_{\text{fc1}})
\end{equation}
where \( h \in \mathbb{R}^{128} \) (if 128 hidden units are used), \( W_{\text{fc1}} \in \mathbb{R}^{128 \times n} \) is the weight matrix, and \( b_{\text{fc1}} \in \mathbb{R}^{128} \) is the bias vector. The ReLU introduces non-linearity, while dropout (with a rate of 0.25) mitigates overfitting. Finally, the output layer applies a sigmoid function to yield probability values for each feature:
\begin{equation}
\hat{q} = \sigma(W_{\text{fc2}} \cdot h + b_{\text{fc2}})
\end{equation}
where \( \hat{q} = (\hat{q}_1, \hat{q}_2, \ldots, \hat{q}_d) \) represents the probability score for each feature. The sigmoid activation ensures that \( \hat{q}_j \in [0, 1] \) for all \( j \), making \( \hat{q} \) interpretable as a relevance score or probability vector and \( \sigma(z) = \frac{1}{1 + e^{-z}} \) is the sigmoid function.

\subsubsection{Distribution-Based Sampling for Probabilistic Feature Selection}
In our framework, feature selection is modeled as a probabilistic policy \( \pi(a \mid \hat{q}) \), where each feature selection mask \( a \in \{0, 1\}^d \) is sampled based on a probability vector \( \hat{q} = (\hat{q}_1, \hat{q}_2, \ldots, \hat{q}_d) \). This vector \( \hat{q} \), derived from the network output, represents the relevance probability of each feature. For a given action \( a \) (or feature mask), the policy is defined as:
\begin{equation}
\pi(a \mid \hat{q}) = \prod_{j=1}^{d} P(a_j \mid \hat{q}_j)
\end{equation}
where \( P(a_j \mid \hat{q}_j) \) denotes the probability of selecting (or discarding) feature \( j \) conditioned on its relevance score \( \hat{q}_j \). This probabilistic formulation introduces flexibility by allowing the network to dynamically control the inclusion probability of each feature through its learned relevance estimate.

To parameterize the feature selection probabilities, we employ the Beta distribution, a continuous distribution bounded within \([0,1]\), which is well-suited for modeling probabilities, as shown in Table~\ref{tab:distribution}. Each feature’s selection probability is sampled as $a_j \sim \text{Beta}(\alpha_j, \beta_j)$, where the shape parameters are computed as:
\begin{equation}
\alpha_j = \beta_j = \text{softplus}(\hat{q}_j) + 1
\end{equation}
Here, \( \text{softplus}(x) = \ln(1 + e^x) \) ensures positivity and numerical stability for the shape parameters. The symmetry \( \alpha_j = \beta_j \) allows the distribution to smoothly vary from uniform to highly skewed based on the learned relevance score \( \hat{q}_j \). When \( \hat{q}_j \) is large, the Beta distribution skews toward 1 (favoring feature retention), and when small, it skews toward 0 (favoring feature exclusion). This approach provides a highly flexible and differentiable mechanism for stochastic feature selection.

The sampled feature mask \( \mathbf{a} \) is then applied to the feature vector \( \mathbf{f} \) as $\mathbf{f}_{\text{selected}} = \mathbf{f} \odot \mathbf{a}$, where \( \odot \) denotes element-wise multiplication, yielding a filtered feature vector \( \mathbf{f}_{\text{selected}} \) that selectively retains features based on their relevance.

\subsubsection{Feedback Alignment with MSE Loss}
FSNet uses mean squared error (MSE) to align the model’s feature relevance scores with human-provided feedback. Let \( f \in \mathbb{R}^d \) represent a feedback vector where each \( f_j \in [0,1] \) indicates human-assessed importance for feature \( j \). The feedback cost function \( C_f \) is defined as:
\begin{equation}\label{eq:feedback}
\mathcal{L}_{\text{feedback}} = C_f(x; \hat{q}, f) = \mathbb{E} \left[ \| f - \hat{q} \|^2 \right] = \frac{1}{d} \sum_{j=1}^d (f_j - \hat{q}_j)^2
\end{equation}
This loss function penalizes discrepancies between the model’s probability vector \( \hat{q} \) and the feedback \( f \). Minimizing \( C_f \) encourages FSNet to produce relevance scores that align with human intuition, resulting in a more interpretable feature selection.

\begin{algorithm}[!ht]
\footnotesize
    \caption{Training Procedure for DDQN Model}
    \label{algo:2}
    \KwIn{Configuration \textit{config}, containing training hyperparameters}
    \KwOut{Trained Q-network, performance metrics}
    Initialize environment $\mathcal{E}$, replay buffer $\mathcal{B}$ with capacity \textit{buffer\_size}\;
    Define $Q_{\theta}$, the Q-network, and $Q_{\theta'}$, the target network\;
    Initialize optimizer with \textit{learning\_rate} and \textit{weight\_decay}\;
    Define the learning rate scheduler with decay factor \textit{$\gamma$}\;

    \For{each epoch $t = 1, 2, \ldots, \textit{n\_epochs}$}{
        Initialize training metrics: running loss, correct predictions, feedback cost\;
        \For{each batch $b$ in environment $\mathcal{E}$}{
            Obtain current state $s$, label $y$, and feedback from environment\;
            \If{\textit{feature\_selection} is True}{
                Apply feature selection using agent, compute feedback cost and probabilities\;
                Store feedback cost and probabilities\;
                Update $s \leftarrow \text{processed state}$\;
            }
            \Else{
                Use raw state\;
            }

            Choose action $a$ using $\epsilon$-greedy on $Q_{\theta}$ or random action if warm-up\;
            Compute reward based on $a$ and $y$\;
            Obtain next state $s'$, apply feature selection if enabled\;
            Store $(s, a, s', r)$ in $\mathcal{B}$\;
            
            \If{buffer $\mathcal{B}$ is ready (size $> \textit{batch\_size}$)}{
                Sample a batch from $\mathcal{B}$\;
                Compute $Q$-learning target $y = r + \gamma \max_{a'} Q_{\theta'}(s', a')$\;
                Compute current estimate $Q_{\theta}(s, a)$\;
                Compute combined loss using $\mathcal{L}_{\text{SmoothL1}}$ with regularization\;
                \If{\textit{feature\_selection} is True}{
                    Add feedback cost to loss\;
                }
                Backpropagate loss and update $Q_{\theta}$\;
            }
        }
        \If{$t \mod 25 = 0$ and \textit{method} is "KAN" and $t < \frac{\textit{n\_epochs}}{2}$}{
            Update grid of $Q_{\theta}$ and $Q_{\theta'}$ with samples from $\mathcal{B}$\;
        }
        
        \If{$t \mod \textit{target\_update} = 0$}{
            Update target $Q_{\theta'} \leftarrow Q_{\theta}$\;
        }
        Adjust learning rate with scheduler\;
    }
    Return $Q_{\theta}$ and collected metrics (accuracy, loss, feedback cost history)\;
\end{algorithm}

\subsection{Double Deep Q-Network Architecture}
The DDQN architecture in our approach extends the traditional Q-learning framework by employing two neural networks: the primary \( Q \)-network, denoted as \( Q_{\theta} \), and a target network, denoted as \( Q_{\theta'} \). The primary \( Q \)-network learns the action-value function \( Q(s, a; \theta) \), estimating the expected cumulative reward for selecting action \( a \) in a given state \( s \). To stabilize training and mitigate the overestimation bias commonly observed in standard Q-learning, we use the target network \( Q_{\theta'} \), updated less frequently than the primary network.

During training, the parameters of \( Q_{\theta} \) are updated via gradient descent, while the parameters of \( Q_{\theta'} \) are periodically synchronized with \( Q_{\theta} \) to avoid target instability. To update \( Q(s, a; \theta) \) toward the target (Equation \ref{eq:target}), we leverage a replay buffer \(\mathcal{B}\), which stores experience tuples \((s, a, r, s')\). Sampling mini-batches from \(\mathcal{B}\) helps reduce correlations between consecutive experiences, improving stability and allowing for independent updates to Q-values.

\subsection{KAN }\label{method:kan}
We now instantiate KAN as the DDQN head, following the background in Sec. {\ref{sec:2}}-{\ref{back:kan}}.In our approach, the KAN-based architecture is structured with widths of [64, 8, 10] for the input, hidden, and output layers. Both the Q-network and target network use this compact configuration to capture complex feature interactions efficiently. Each KAN layer combines a spline function and a residual basis function, formulated as:
\begin{equation}
\phi(x) = w_b b(x) + w_s spline(x)
\end{equation}
where \(b(x) = silu(x)\) serves as a non-linear basis function, while \(spline(x) = \sum_i c_i B_i(x)\) utilizes trainable B-splines for flexible approximation. With spline order \(k=3\) over a specified grid, this layered structure dynamically adjusts to feature relevance throughout training. For optimization, \(w_b\) and \(w_s\) are initialized carefully: \(w_b\) uses Xavier initialization, ensuring balanced layer activation scales, while \(w_s\) is set to 1, with the initial spline function close to zero. The model’s spline grids adapt to changing input activations, extending KAN’s effective region for learning. The choice of KAN for the Q-network is theoretically motivated by its property of local plasticity, which mitigates catastrophic forgetting common in global MLP updates. We provide a detailed theoretical analysis of KAN approximation bias and gradient stability in Appendix \ref{app:kan}.

\subsection{MLP}
The MLP-based architecture provides a more traditional setup for the DDQN framework, using a straightforward, fully connected design for both the Q-network and the target network. This configuration comprises two linear layers: a 64-neuron initial layer that maps the input features to a hidden space defined by a 32-neuron network width, followed by a ReLU activation, and a 10-neuron final linear layer that produces Q-values. The MLP requires a larger hidden layer with 32 neurons to approximate the complex mappings found in datasets, making it parameter-heavy relative to the KAN-based DDQN.

\subsection{Training Procedure: KAN-based and MLP-based DDQN}
The training procedure for the DDQN is illustrated in Algorithm \ref{algo:2} and Fig. \ref{fig:system}, which outlines the main steps taken during each epoch. The algorithm initializes the environment and Q-networks and iteratively processes batches of experiences from a replay buffer \( \mathcal{B} \). Each experience consists of the current state \( s \), action \( a \), reward \( r \), and the next state \( s' \).

We employ two training algorithms to optimize our Q-network, each tailored to different model architectures: (1) a KAN-based training procedure that utilizes regularization for interpretability and stability, and (2) a standard MLP-based training procedure. Each training procedure integrates feedback cost when feature selection is enabled, promoting a minimalistic and interpretable feature set.

\subsubsection{Temporal-Difference Target Computation}
Both KAN and MLP-based approaches utilize the temporal-difference (TD) target to stabilize Q-learning updates and reduce the discrepancy between estimated Q-values and TD targets. For a given experience tuple \((s, a, s', r, d)\), where \( s \) and \( s' \) represent the current and next states, \( a \) is the action taken, \( r \) is the reward, and \( d \) indicates the termination flag, the TD target is computed as:
\begin{equation}\label{eq:target}
TD_{target} = r + \gamma \, Q_{\theta'}\left(s', \arg\max_{a} Q_{\theta}(s', a)\right)
\end{equation}
where \( \gamma \in [0, 1] \) is the discount factor for future rewards. Here, \( Q_{\theta}(s', a) \) determines the action \( a \) that maximizes the Q-value for the next state \( s' \), and \( Q_{\theta'} \) provides the stable estimate for this chosen action. The target network \( Q_{\theta'} \), updated less frequently, offers a stable estimation for TD updates.

\subsubsection{KAN-based Training with Regularization}
In the KAN-based training procedure, we employ $Smooth L1$ loss, which has proven effective in mitigating the influence of large outliers in Q-value errors. The primary loss term \( \mathcal{L}_{\text{TD}} \) is given by:
\begin{equation}
\mathcal{L}_{\text{TD}} = L_{\delta}\left(Q_{\theta}(s, a) - TD_{target}\right)
\end{equation}
where \( L_{\delta} \) represents the $Smooth L1$ loss:
\begin{equation}
L_{\delta}(x) = 
\begin{cases} 
0.5x^2 & \text{if } |x| < 1 \\
|x| - 0.5 & \text{otherwise}
\end{cases}
\end{equation}

\textit{L1 and Entropy-Based Regularization:} In KANs, the L1 norm is applied to encourage sparsity in the activation functions, which replaces traditional linear weights used in MLPs. The L1 norm of an activation function \( \phi \) is defined as the average magnitude of its outputs over \( N_p \) inputs:
\begin{equation}
\|\phi\|_1 \equiv \frac{1}{N_p} \sum_{s=1}^{N_p} \phi(x^{(s)})
\end{equation}
For a KAN layer \( \Phi \) with \( n_{in} \) inputs and \( n_{out} \) outputs, we define the L1 norm of \( \Phi \) as the sum of the L1 norms of all activation functions:
\begin{equation}
|\Phi|_1 \equiv \sum_{i=1}^{n_{in}} \sum_{j=1}^{n_{out}} \|\phi_{i,j}\|_1.
\end{equation}
Additionally, an entropy term is introduced to mitigate overconfidence in the predictions. The entropy \( S(\Phi) \) for the KAN layer is given by:
\begin{equation}
S(\Phi) \equiv -\sum_{i=1}^{n_{in}} \sum_{j=1}^{n_{out}} \frac{|\phi_{i,j}|_1}{|\Phi|_1} \log\left(\frac{|\phi_{i,j}|_1}{|\Phi|_1}\right)
\end{equation}
The combined regularization term is defined as:
\begin{equation}
\mathcal{R}_{\text{L1+Entropy}} = \sum_{i=1}^{n} \left( \lambda_{\text{L1}} \|\textit{acts\_scale}_i\|_1 - \lambda_{\text{entropy}} H(\textit{acts\_scale}_i) \right)
\end{equation}
where \(H(\textit{acts\_scale}_i) = - \sum p \log(p)\) denotes entropy with \(p\) being the normalized activation values, and \( \lambda_{\text{L1}} \) and \( \lambda_{\text{entropy}} \) are hyperparameters.

\textit{Spline-Based Regularization:} This term regularizes the spline coefficients of the KAN activation functions, encouraging smooth transitions and sparsity in the feature space. Given the spline coefficient vector \(\text{coef}_i\) of each activation function, we compute the following:
\begin{equation}
\mathcal{R}_{\text{Spline}} = \sum_{i=1}^{n} \left( \lambda_{\textit{coef}} \|\textit{coef}_i\|_1 + \lambda_{\textit{coefdiff}} \|\textit{diff}(\textit{coef}_i)\|_1 \right)
\end{equation}
where \(\textit{diff}(\textit{coef}_i)\) calculates the adjacent differences within \(\textit{coef}_i\), enforcing smoothness in the function and preventing rapid oscillations in the learned coefficients. The combined regularization term \( \mathcal{R} \) is added to the loss as follows:

\footnotesize
\begin{equation}
\mathcal{L}_{\text{KAN}} = \mathcal{L}_{\text{TD}} + \lambda \mathcal{R} = L_{\delta}\left(Q_{\theta}(s, a), TD_{target}\right) + \lambda \left( \mathcal{R}_{\text{L1+Entropy}} + \mathcal{R}_{\text{Spline}} \right)
\end{equation}\normalsize

\subsubsection{Feedback Cost Integration for Feature Selection}
To minimize unnecessary feature dependencies, we incorporate a feedback cost \(\mathcal{L}_{\text{feedback}}\) (Equation \ref{eq:feedback}) into the total loss with an $\alpha$ = 0.5 when feature selection is enabled. This cost penalizes selected features based on their contribution to the model, defined as follows:
\begin{equation}\label{eq20}
\mathcal{L}_{\text{KAN, total}} = \mathcal{L}_{\text{KAN}} + \alpha \times \mathcal{L}_{\text{feedback}}
\end{equation}

\subsubsection{MLP-based Training}
For the MLP-based training, we use the same TD target computation and $Smooth L1$ loss but omit the KAN-specific regularization terms. Like KAN-based training, feature selection introduces an additional feedback cost term to this loss: $\mathcal{L}_{\text{MLP, total}} = L_{\delta}\left(Q_{\theta}(s, a), TD_{target}\right) + \alpha \times \mathcal{L}_{\text{feedback}}$.

\subsubsection{Optimization and Gradient Clipping}
To ensure stable training and mitigate the risk of exploding gradients, we apply in-place gradient clipping with a threshold of 100 to both KAN and MLP-based training. This process limits the magnitude of gradients, preventing extreme updates that could destabilize the learning process: $\textit{clip}\left(\frac{\partial \mathcal{L}}{\partial \theta}, 100\right)$. The gradients are then backpropagated, and the optimizer updates the network weights to minimize \(\mathcal{L}_{\text{KAN, total}}\) in KAN and \(\mathcal{L}_{\text{MLP, total}}\) in MLP model.

\subsection{Integration of HITL Feedback in DDQN}
Integrating HITL feedback into the DDQN architecture enables the model to iteratively improve its feature selection by incorporating human expertise. This process adjusts the feature selection probabilities  \( \hat{q}_j \) based on human feedback, leading to an improved and rationale-aware selection policy.

\subsubsection{Feature Selection Adjustment Using HITL Feedback}
For each feature \( j \), HITL feedback provides a target value \( f_j \), representing the importance of that feature according to human assessment. The DDQN uses this feedback to adjust the selected feature probabilities by minimizing the feedback cost \( \mathcal{L}_{\text{feedback}} \), thereby aligning the DDQN’s policy with human expertise.

\subsubsection{Action-Selection Policy with HITL Feedback}
The DDQN framework selects actions via an \(\epsilon\)-greedy policy, which balances exploration and exploitation. At the start of training, \(\epsilon\) is set to a high value, allowing the agent to explore actions and discover potentially valuable states randomly. Over time, \(\epsilon\) gradually decays, encouraging the agent to exploit its learned policy by selecting actions that maximize the estimated Q-value from the primary \( Q_{\theta} \) network:
\begin{equation}
a = \begin{cases} 
\text{random action} & \text{with probability } \epsilon \\
\arg\max_{a'} Q(s, a'; \theta) & \text{with probability } 1 - \epsilon 
\end{cases}
\end{equation}
This approach improves learning by promoting diverse experiences early on while progressively focusing on reliable, high-reward actions as training progresses. When HITL feedback is introduced, the DDQN updates its policy to emphasize features positively reinforced by feedback, using this guidance to improve feature selection iteratively and refine the policy across epochs. Therefore, this \(\epsilon\)-greedy mechanism prevents premature convergence to suboptimal policies, supporting more robust training for effective feature selection.

\subsubsection{Iterative Learning Process}
The training loop for DDQN, with HITL feedback, ensures that the Q-network \( Q_{\theta} \) converges to an optimal feature selection policy. At each epoch, the Q-network’s weights are updated based on both the Q-learning target and the feedback cost. This iterative process continues as:
\begin{equation}
\theta \leftarrow \theta - \eta \nabla_{\theta} \mathcal{L}_{\text{total}}
\end{equation}
where \( \eta \) is the learning rate and \( \alpha \) is a hyperparameter balancing the feedback cost and Q-network loss. By adjusting \( \alpha \) over time, the DDQN increasingly aligns its feature selection with human insights, resulting in a more interpretable and effective feature set.

\subsubsection{Practical Implications and Robustness}
While our framework utilizes simulated Gaussian feedback to model ideal expert attention, we discuss the practical implications, robustness to noisy annotations, and sensitivity to the loss function in Appendix \ref{app:hitl_robustness}.

\section{Experimental Design}\label{sec:4}

\subsection{Dataset Preparation}
We used a benchmark environment to standardize data preparation, feedback simulation, and visualizations across four datasets: MNIST, FashionMNIST, CIFAR-10, and CIFAR-100. Each dataset was resized to a standardized input dimension of \(8 \times 8\) pixels, with normalization applied based on empirically computed means and standard deviations specific to each dataset to aid in training convergence. For MNIST, we used a mean ($\mu$) of 0.1307 and a standard deviation ($\sigma_{SD}$) of 0.3081; for FashionMNIST, both $\mu$ and $\sigma_{SD}$ were set to 0.5. For CIFAR-10, normalization employed channel-wise $\mu$ of $(0.4914, 0.4822, 0.4465)$ and $\sigma_{SD}$ of $(0.247, 0.243, 0.261)$, while CIFAR-100 used $(0.5071, 0.4865, 0.4409)$ and $(0.2673, 0.2564, 0.2761)$ respectively. This normalization process adjusts pixel intensity \(\mathbf{I}\) via the formula:
\begin{equation}
\mathbf{I}_{\text{norm}} = \frac{\mathbf{I} - \mu}{\sigma_{SD}}
\end{equation}
where \(\mu\) and \(\sigma_{SD}\) are dataset-specific values.

Batch processing was performed with a batch size of \( B = 128 \), enabling efficient data handling and parallel processing. Once a batch reaches its end, the iterator is reset, ensuring a continuous data stream throughout training. This setup streamlined the flow of images into memory along with the generated feedback signals for each batch.

\begin{figure}[!ht]
\centering
\includegraphics[width=1\linewidth]{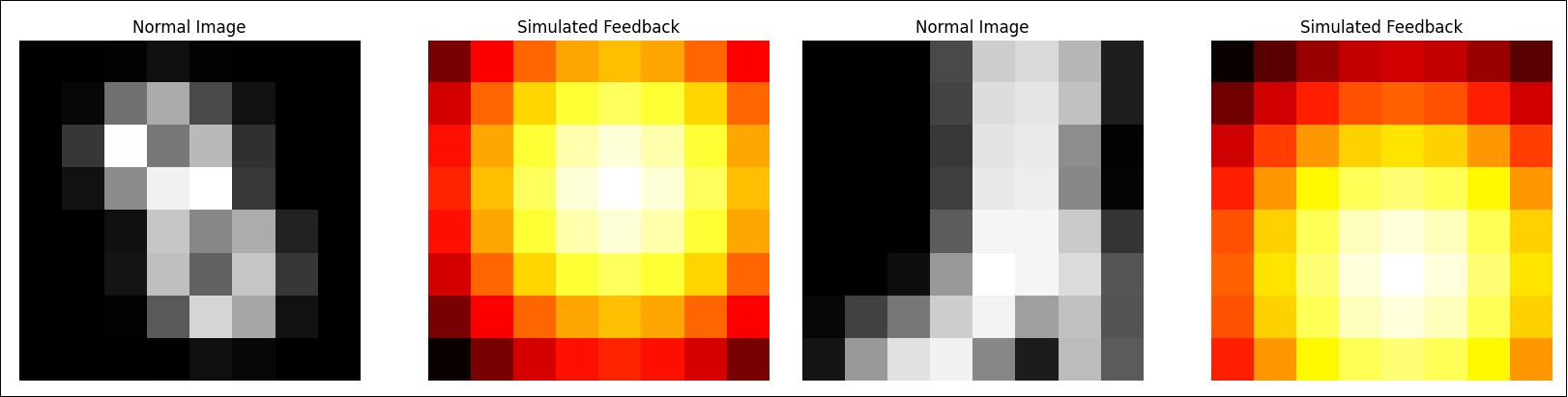}
\caption{A sample of preprocessed (\(8 \times 8\)) pixel MNIST (left) and FashionMNIST (right) images and their corresponding feedback maps (\(\sigma = 5.0\)).}
\label{fig:feedback}
\end{figure}


\subsection{Simulated Feedback Mechanism}
To simulate supervisory guidance, we designed a feedback mechanism that highlights the most salient regions within each image via a Gaussian heatmap. The process begins with normalizing the input image \( I \in \mathbb{R}^{C \times H \times W} \) to a range of \([0, 1]\), defined by the transformation \( I_{\text{norm}} = \frac{I - \min(I)}{\max(I) - \min(I)} \). This normalization ensures that all pixel values are scaled appropriately for subsequent computations. Next, the center of mass \( C \) of the image is identified, represented by the coordinates \((x_c, y_c)\) corresponding to the pixel with the maximum intensity, calculated using \( C = \text{argmax}(I_{\text{norm}}) \). Utilizing these coordinates, a Gaussian feedback mask \( M(x, y) \) is generated, which is centered around the identified pixel. The equation defines the Gaussian mask:
\begin{equation}
M(x, y) = \exp\left(-\frac{(x - x_c)^2 + (y - y_c)^2}{2 \eta^2}\right)
\end{equation}
where \( \eta = 5.0\) is a parameter that controls the spread of the Gaussian distribution, reflecting the degree of uncertainty in the feedback. This mask provides spatial feedback, where values decay radially from the center (see Figure \ref{fig:feedback}). To ensure that the feedback mask is interpretable and usable within the model, it is normalized so that its maximum value equals 1: $M_{\text{norm}} = \frac{M}{\max(M)}$. The resulting normalized feedback mask effectively emphasizes regions within the image with higher intensity values, which aligns to guide the model's attention toward salient features.

\subsection{Resources Used}
Our experiments were conducted via a setup powered by an Intel(R) Xeon(R) CPU with a clock speed of 2 GHz, 4 virtual CPU cores, and 16 GB of DDR4 RAM. The software environment included Python 3.10.12, with PyTorch for model development, PyKAN for KAN, Gymnasium for environment simulations, SciPy and NumPy for numerical computations, Matplotlib for data visualization, and SymPy for symbolic calculations.

\subsection{Experimental Configurations}
For the experiments conducted in this study, we employed two configurations for the MLP and KAN models, both set to a batch size of 128 and trained over 100 epochs. The configurations utilized a learning rate $1\times10^{-3}$, a weight decay $1\times10^{-4}$, and a discount factor ($\gamma$) of 0.99. The MLP configuration featured a width of 32 and an output size of 10 classes, while the KAN configuration used a width of 8 and a grid size of 3. Both models share a common input size of 64 features, a buffer size of 100,000, and a target update interval of 10. They incorporated warm-up episodes of 2 and utilized a beta distribution (see Table \ref{tab:distribution}) for their stochastic feature selection processes, with an initial $\tau$ value set at 1.0. We strategically designed the architecture of our model by selecting the appropriate combination of convolutional layers and filters to ensure that the resulting selected feature map retains the same shape as the input and feedback size (64 for \(8 \times 8\) images). This choice is crucial, as it directly allows the simulated feedback (similar to input size) to correspond to the feature map dimensions, facilitating effective integration during the feature selection process. In this case, for 1, 2, and 3 convolutional layers, we must select 2, 4, and 8 filter sizes to maintain the desired selected feature shape. Detailed descriptions of the evaluation metrics (accuracy, calibration, sparsity, and complexity), training of baselines, ablation protocols, and visualization scope are provided in Appendices~\ref{app:eval}, \ref{app:baseline}, \ref{app:ablation}, and \ref{app:scope}.

\begin{figure}[!ht]
\centering
\subfloat[\label{accuracy1a}]{
\includegraphics[width=1\linewidth]{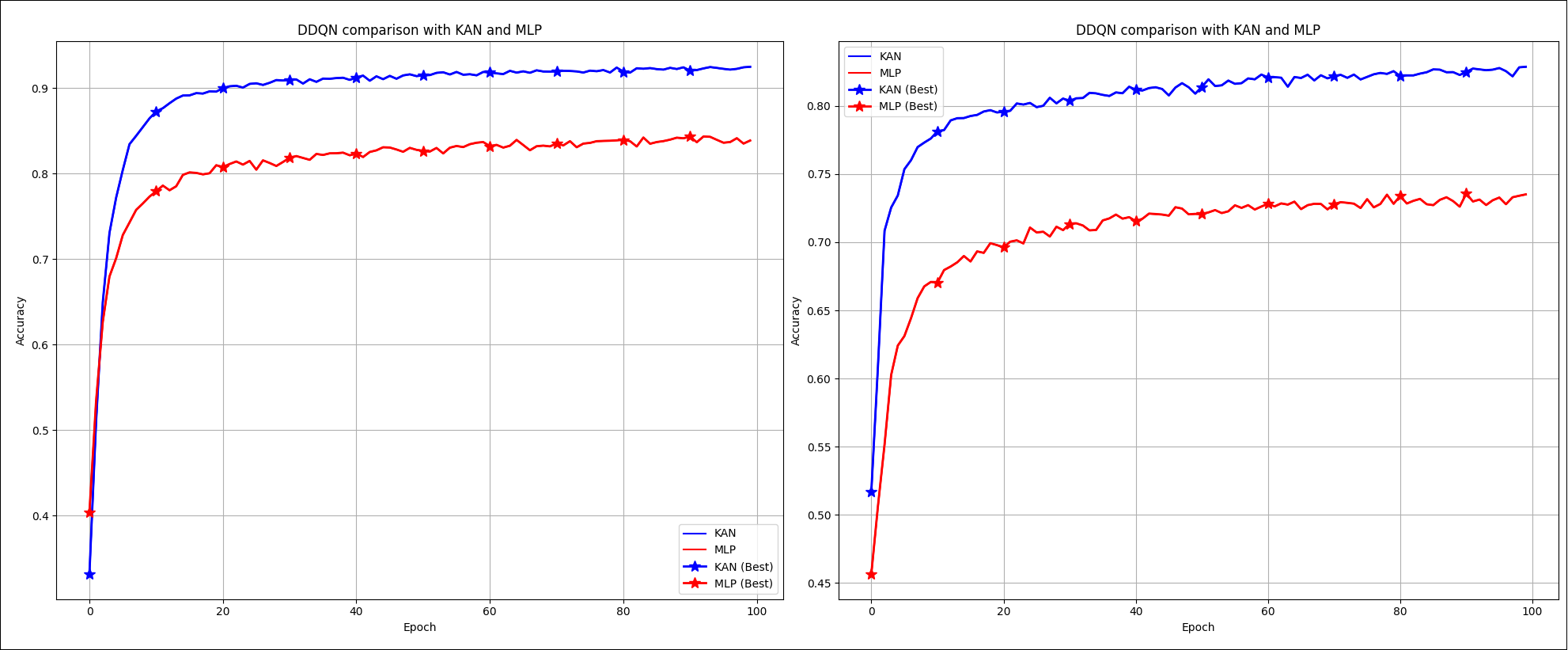}
}
\hfill
\subfloat[\label{accuracy1b}]{
\includegraphics[width=1\linewidth]{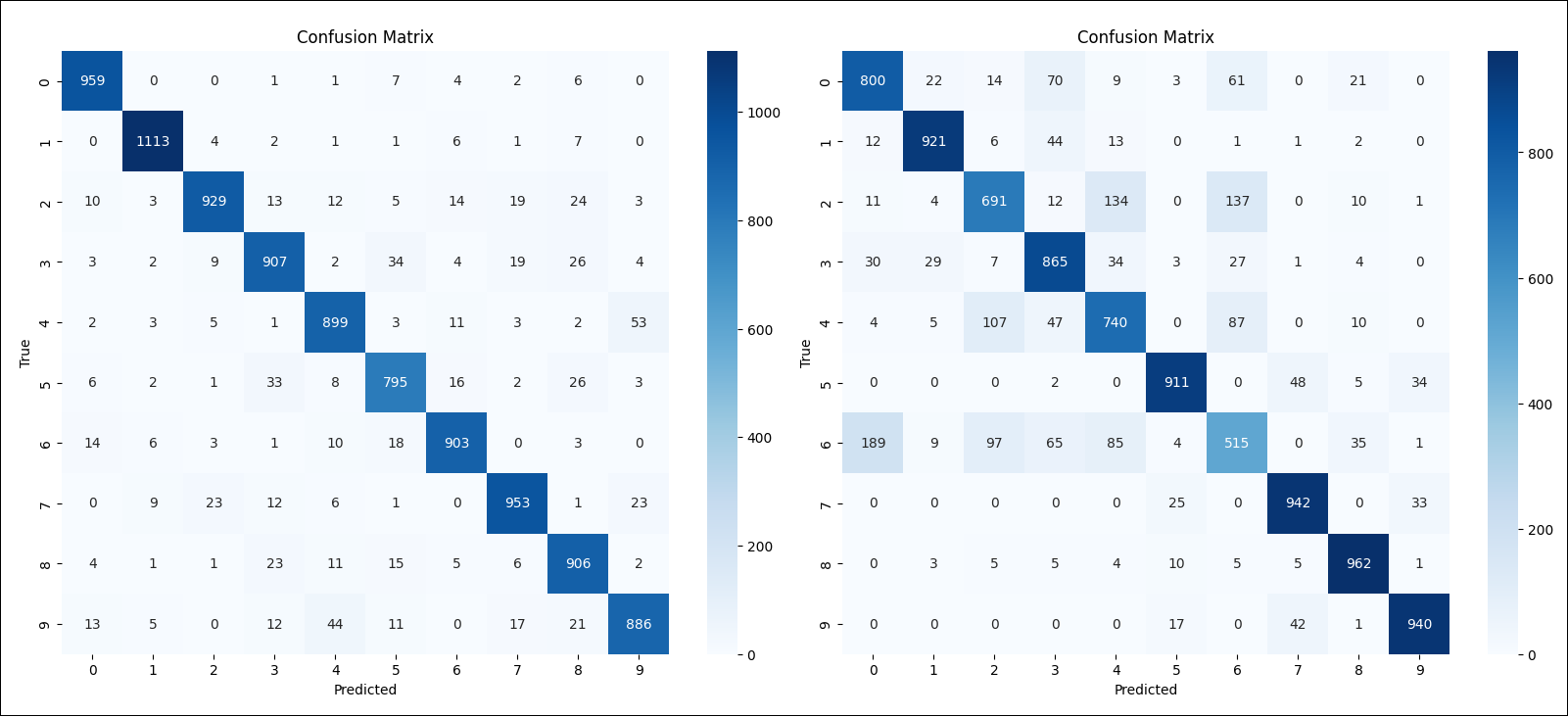}
}
\caption{(a) Training accuracy per epoch comparison of KAN and MLP-based DDQN on MNIST (left) and FashionMNIST (right), and (b) confusion matrix of KAN-DDQN on MNIST (left) and FashionMNIST (right).}
\label{fig:accuracy}
\end{figure}

\begin{figure}[!ht]
\centering
\subfloat[\label{interpret1a}]{
\includegraphics[width=0.45\linewidth]{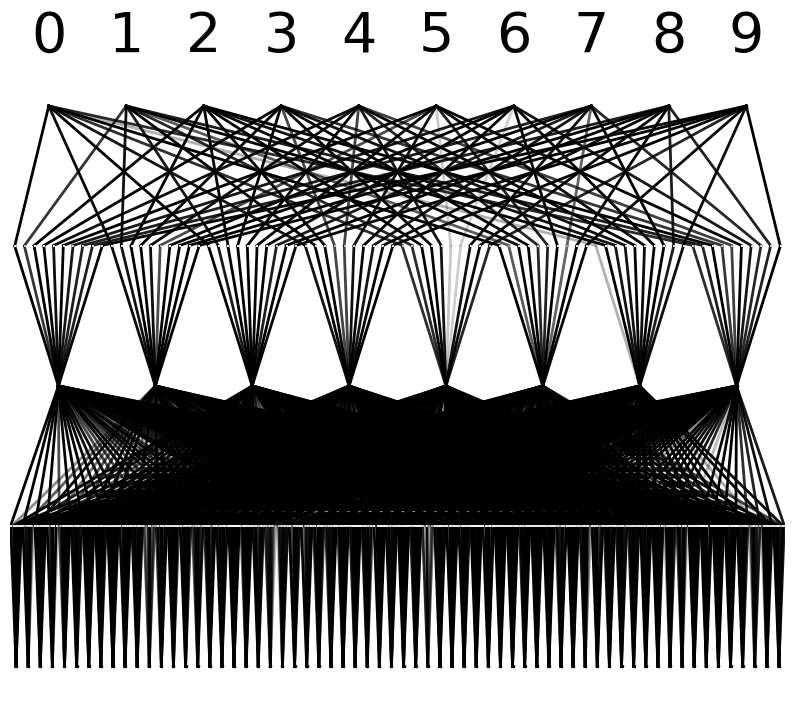}
}
\hfill
\subfloat[\label{interpret1b}]{
\includegraphics[width=0.45\linewidth]{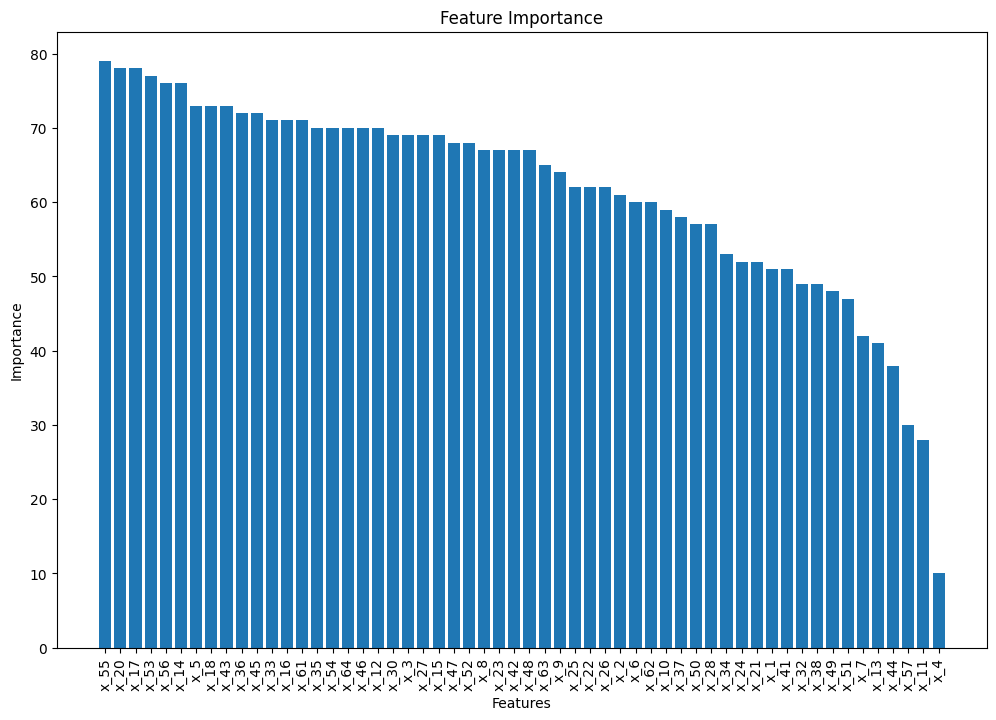}
}
\hfill
\subfloat[\label{interpret1c}]{
\includegraphics[width=1\linewidth]{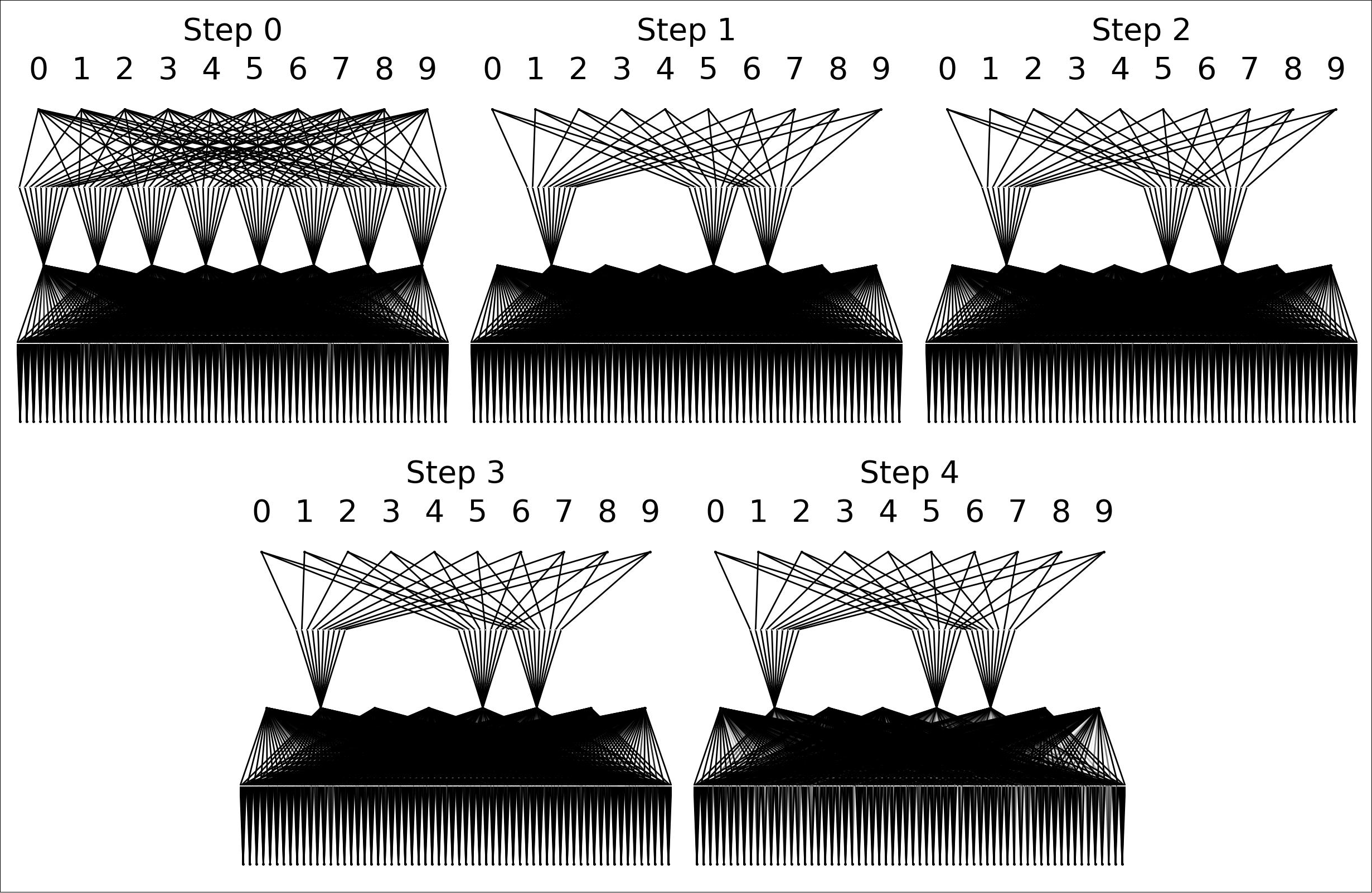}
\hfill
}
\caption{(a) Pruned KAN architecture after removing neurons with both low incoming and outgoing activation magnitudes, exposing a minimal functional skeleton; (b) feature importances extracted from auto-symbolic forms of learned splines, indicating how often input features appear in simplified expressions; (c) early training trajectory (first five steps), where the policy concentrates mass on discriminative inputs before spreading to additional features. Together, these link sparsity, symbolic structure, and policy behavior.}
\label{fig:interpret1}
\end{figure}

\begin{figure}[!ht]
\centering
\includegraphics[width=1\linewidth]{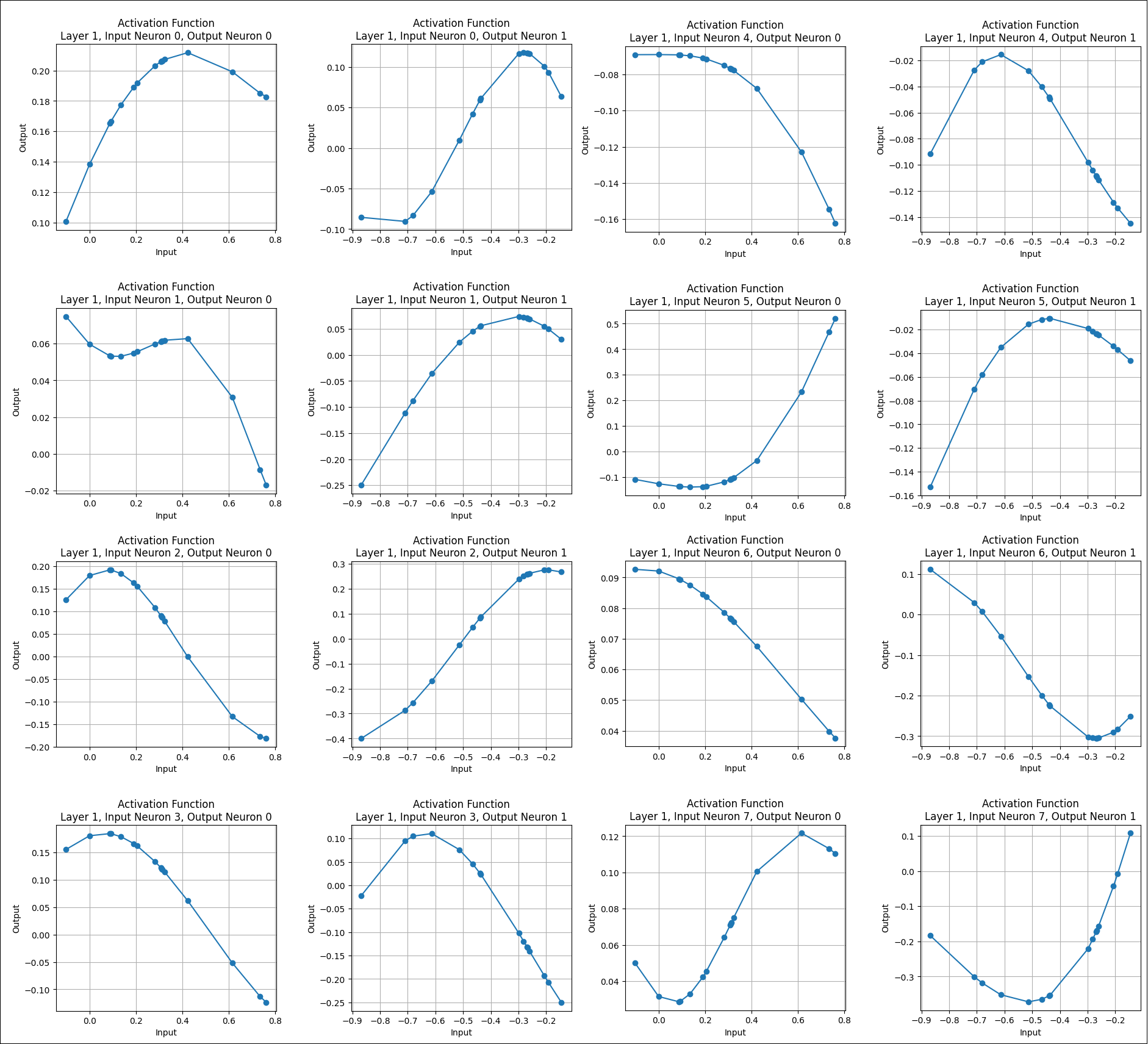}
\caption{Activation functions for middle neurons in the KAN agent's hidden layer on the MNIST dataset. Each subplot shows how a specific neuron transforms inputs via the spline activation, illustrating various activation behaviors across the layer.}
\label{fig:active-neurons}
\end{figure}

\begin{table}[!ht]
\centering
\caption{Performance of MLP-DDQN for different distributions on MNIST and FashionMNIST datasets. The $\mu$ and $\sigma_{SD}$ are computed over a 10-epoch moving window. \textbf{Bold} indicates the best performance.}
\label{tab:distribution}
\resizebox{\columnwidth}{!}{%
\begin{tabular}{lcccc}
\toprule
\multirow{2}{*}{\textbf{Distribution}} &
  \multicolumn{2}{c}{\textbf{MNIST}} &
  \multicolumn{2}{c}{\textbf{FashionMNIST}} \\ \cline{2-5} 
 &
  \multicolumn{1}{c}{\textbf{Train Accuracy}} &
  \textbf{Test Accuracy} &
  \multicolumn{1}{c}{\textbf{Train Accuracy}} &
  \textbf{Test Accuracy} \\ \midrule
Bernoulli      & \multicolumn{1}{c}{0.57 $\pm$ 0.03} & 0.58 $\pm$ 0.00 & \multicolumn{1}{c}{0.55 $\pm$ 0.02} & 0.54 $\pm$ 0.00 \\ 
Gumbel Softmax & \multicolumn{1}{c}{0.15 $\pm$ 0.01} & 0.18 $\pm$ 0.00 & \multicolumn{1}{c}{0.10 $\pm$ 0.02} & 0.24 $\pm$ 0.00 \\ 
Gaussian       & \multicolumn{1}{c}{0.22 $\pm$ 0.01} & 0.21 $\pm$ 0.00 & \multicolumn{1}{c}{0.22 $\pm$ 0.01} & 0.21 $\pm$ 0.01 \\ 
\textbf{Beta} &
  \multicolumn{1}{c}{\textbf{0.79 $\pm$ 0.02}} &
  \textbf{0.78 $\pm$ 0.00} &
  \multicolumn{1}{c}{\textbf{0.68 $\pm$ 0.01}} &
  \textbf{0.66 $\pm$ 0.00} \\ 
Dirichlet      & \multicolumn{1}{c}{0.17 $\pm$ 0.02} & 0.22 $\pm$ 0.00 & \multicolumn{1}{c}{0.23 $\pm$ 0.02} & 0.24 $\pm$ 0.00 \\ 
Multinomial    & \multicolumn{1}{c}{0.10 $\pm$ 0.02} & 0.12 $\pm$ 0.00 & \multicolumn{1}{c}{0.10 $\pm$ 0.01} & 0.12 $\pm$ 0.00 \\ 
Cauchy         & \multicolumn{1}{c}{0.12 $\pm$ 0.01} & 0.12 $\pm$ 0.00 & \multicolumn{1}{c}{0.11 $\pm$ 0.01} & 0.12 $\pm$ 0.00 \\ 
Laplace        & \multicolumn{1}{c}{0.16 $\pm$ 0.02} & 0.18 $\pm$ 0.00 & \multicolumn{1}{c}{0.21 $\pm$ 0.01} & 0.19 $\pm$ 0.00 \\ 
Uniform        & \multicolumn{1}{c}{0.73 $\pm$ 0.02} & 0.75 $\pm$ 0.00 & \multicolumn{1}{c}{0.69 $\pm$ 0.03} & 0.65 $\pm$ 0.00 \\ \bottomrule
\end{tabular}%
}
\end{table}

\begin{table}[!ht]
\centering
\footnotesize
\caption{Performance Metrics without Feature Selection on MNIST and FashionMNIST Datasets Using KAN and MLP Models.}
\label{tab:no-fs}
\resizebox{\linewidth}{!}{%
\begin{tabular}{llcccccc}
\toprule
\multirow{2}{*}{\textbf{Model}} &
  \multirow{2}{*}{\textbf{Dataset}} &
  \multicolumn{4}{c}{\textbf{No Feature Selection}} &
  \multirow{2}{*}{\textbf{Training Time (hh:mm:ss)}} \\ \cline{3-6}
 &
   &
  \multicolumn{1}{c}{\textbf{maP}} &
  \multicolumn{1}{c}{\textbf{maR}} &
  \multicolumn{1}{c}{\textbf{maF1}} &
  \textbf{Test Accuracy} &
   \\ \midrule
KAN & MNIST        & \multicolumn{1}{c}{0.57} & \multicolumn{1}{c}{0.57} & \multicolumn{1}{c}{0.56} & \textbf{0.58 $\pm$ 0.01} & 5:10:50 \\ 
MLP & MNIST        & \multicolumn{1}{c}{0.45} & \multicolumn{1}{c}{0.51} & \multicolumn{1}{c}{0.47} & 0.52 $\pm$ 0.00 & 10:41   \\ 
KAN & FashionMNIST & \multicolumn{1}{c}{0.63} & \multicolumn{1}{c}{0.65} & \multicolumn{1}{c}{0.63} & \textbf{0.64 $\pm$ 0.00} & 5:10:32 \\ 
MLP & FashionMNIST & \multicolumn{1}{c}{0.55} & \multicolumn{1}{c}{0.60}  & \multicolumn{1}{c}{0.57} & 0.59 $\pm$ 0.00 & 10:24   \\ \bottomrule
\end{tabular}%
}
\end{table}

\begin{table*}[!ht]
\centering
\caption{Impact of FSNet parameters on test accuracy using MNIST and FashionMNIST datasets with feature selection}
\label{tab:with-fs}
\resizebox{\textwidth}{!}{%
\begin{tabular}{llcccccccccc}
\toprule
\multirow{2}{*}{\textbf{Model}} &
  \multirow{2}{*}{\textbf{Dataset}} &
  \multirow{2}{*}{\textbf{\# Conv Layers}} &
  \multirow{2}{*}{\textbf{\# Filters}} &
  \multirow{2}{*}{\textbf{\# Width}} &
  \multirow{2}{*}{\textbf{Grid Size}} &
  \multicolumn{5}{c}{\textbf{With Feature Selection}} &
  \multirow{2}{*}{\textbf{Training Time (hh:mm:ss)}} \\ \cline{7-11}
 &
   &
   &
   &
   &
   &
  \multicolumn{1}{c}{\textbf{maP}} &
  \multicolumn{1}{c}{\textbf{maR}} &
  \multicolumn{1}{c}{\textbf{maF1}} &
  \multicolumn{1}{c}{\textbf{Train Accuracy}} &
  \textbf{Test Accuracy} &
   \\ \midrule
KAN &
  MNIST &
  \textbf{1} &
  \textbf{2} &
  8 &
  3 &
  \multicolumn{1}{c}{0.92} &
  \multicolumn{1}{c}{0.92} &
  \multicolumn{1}{c}{0.92} &
  \multicolumn{1}{c}{0.93 $\pm$ 0.01} &
  \textbf{0.93 $\pm$ 0.00} &
  4:35:26 \\ 
MLP &
  MNIST &
  1 &
  2 &
  32&
  - &
  \multicolumn{1}{c}{0.84} &
  \multicolumn{1}{c}{0.84} &
  \multicolumn{1}{c}{0.84} &
  \multicolumn{1}{c}{0.85 $\pm$ 0.02} &
  0.84 $\pm$ 0.00 &
  8:42 \\ 
KAN &
  FashionMNIST &
  \textbf{1} &
  \textbf{2} &
  8 &
  3 &
  \multicolumn{1}{c}{0.83} &
  \multicolumn{1}{c}{0.83} &
  \multicolumn{1}{c}{0.83} &
  \multicolumn{1}{c}{0.85 $\pm$ 0.01} &
  \textbf{0.83 $\pm$ 0.00} &
  4:25:41 \\ 
MLP &
  FashionMNIST &
  1 &
  2 &
  32&
  - &
  \multicolumn{1}{c}{0.72} &
  \multicolumn{1}{c}{0.73} &
  \multicolumn{1}{c}{0.72} &
  \multicolumn{1}{c}{0.73 $\pm$ 0.01} &
  0.74 $\pm$ 0.00 &
  8:17 \\ 
KAN &
  MNIST &
  2 &
  4 &
  8 &
  3 &
  \multicolumn{1}{c}{0.73} &
  \multicolumn{1}{c}{0.73} &
  \multicolumn{1}{c}{0.73} &
  \multicolumn{1}{c}{0.75 $\pm$ 0.02} &
  0.73 $\pm$ 0.00 &
  5:50:29 \\ 
MLP &
  MNIST &
  2 &
  4 &
  32&
  - &
  \multicolumn{1}{c}{0.72} &
  \multicolumn{1}{c}{0.72} &
  \multicolumn{1}{c}{0.72} &
  \multicolumn{1}{c}{0.75 $\pm$ 0.02} &
  0.72 $\pm$ 0.00 &
  10:32 \\ 
KAN &
  FashionMNIST &
  2 &
  4 &
  8 &
  3 &
  \multicolumn{1}{c}{0.69} &
  \multicolumn{1}{c}{0.7} &
  \multicolumn{1}{c}{0.69} &
  \multicolumn{1}{c}{0.71 $\pm$ 0.01} &
  0.70 $\pm$ 0.00 &
  6:02:45 \\ 
MLP &
  FashionMNIST &
  2 &
  4 &
  32&
  - &
  \multicolumn{1}{c}{0.62} &
  \multicolumn{1}{c}{0.68} &
  \multicolumn{1}{c}{0.64} &
  \multicolumn{1}{c}{0.71 $\pm$ 0.02} &
  0.68 $\pm$ 0.00 &
  11:42 \\ \bottomrule
\end{tabular}%
}
\end{table*}

\begin{table*}[!ht]
\centering
\caption{Impact of KAN Parameter Adjustments on MNIST and FashionMNIST Performance}
\label{tab:effect-grid}
\resizebox{\textwidth}{!}{%
\begin{tabular}{llcccccccccc}
\toprule
\multirow{2}{*}{\textbf{Model}} &
  \multirow{2}{*}{\textbf{Dataset}} &
  \multirow{2}{*}{\textbf{\# Conv Layers}} &
  \multirow{2}{*}{\textbf{\# Filters}} &
  \multirow{2}{*}{\textbf{\# Width}} &
  \multirow{2}{*}{\textbf{Grid Size}} &
  \multicolumn{5}{c}{\textbf{With Feature Selection}} &
  \multirow{2}{*}{\textbf{Training Time (hh:mm:ss)}} \\ \cline{7-11}
 &
   &
   &
   &
   &
   &
  \multicolumn{1}{c}{\textbf{maP}} &
  \multicolumn{1}{c}{\textbf{maR}} &
  \multicolumn{1}{c}{\textbf{maF1}} &
  \multicolumn{1}{c}{\textbf{Train Accuracy}} &
  \textbf{Test Accuracy} &
   \\ \midrule
KAN &
  MNIST &
  2 &
  4 &
  8 &
  2 &
  \multicolumn{1}{c}{0.72} &
  \multicolumn{1}{c}{0.72} &
  \multicolumn{1}{c}{0.72} &
  \multicolumn{1}{c}{0.72 $\pm$ 0.02} &
  0.72 $\pm$ 0.00 &
  6:05:24 \\ 
 KAN & MNIST & 2 & 4 & 8 & 3 & 
 \multicolumn{1}{c}{0.73} & 
 \multicolumn{1}{c}{0.73} & 
 \multicolumn{1}{c}{0.73} & 
 \multicolumn{1}{c}{0.75 $\pm$ 0.02} & 
 0.73 $\pm$ 0.00 &5:50:29 \\ 
KAN &
  MNIST &
  2 &
  4 &
  \textbf{16} &
  \textbf{3} &
  \multicolumn{1}{c}{0.75} &
  \multicolumn{1}{c}{0.74} &
  \multicolumn{1}{c}{0.74} &
  \multicolumn{1}{c}{0.74 $\pm$ 0.02} &
  \textbf{0.75 $\pm$ 0.00} &
  10:21:59 \\ 
KAN &
  FashionMNIST &
  2 &
  4 &
  8 &
  2 &
  \multicolumn{1}{c}{0.70} &
  \multicolumn{1}{c}{0.71} &
  \multicolumn{1}{c}{0.70} &
  \multicolumn{1}{c}{0.72 $\pm$ 0.02} &
  0.71 $\pm$ 0.00 &
  5:11:51 \\ 
 KAN & FashionMNIST & 2 & 4 & 8 & 3 & 
 \multicolumn{1}{c}{0.69} & 
 \multicolumn{1}{c}{0.70} & 
 \multicolumn{1}{c}{0.69} & 
 \multicolumn{1}{c}{0.71 $\pm$ 0.01} & 
 0.70 $\pm$ 0.00 &6:02:45 \\ 
KAN &
  FashionMNIST &
  2 &
  4 &
  \textbf{16} &
  \textbf{3} &
  \multicolumn{1}{c}{0.71} &
  \multicolumn{1}{c}{0.72} &
  \multicolumn{1}{c}{0.71} &
  \multicolumn{1}{c}{0.73 $\pm$ 0.02} &
  \textbf{0.72 $\pm$ 0.00} &
  11:41:30 \\ \bottomrule
\end{tabular}%
}
\end{table*}

\section{Results}\label{sec:5}
Our results reveal that KAN-DDQN outperforms MLP-DDQN in all cases on both MNIST and FashionMNIST datasets (see Fig. \ref{accuracy1a} and \ref{accuracy1b}) with 4 times fewer parameters and only $8\times8$ sized images. Beyond top-1 accuracy, we report macro F1, macro AUC-PR, and calibration (ECE, NLL, Brier, with reliability diagrams) in Appendix~\ref{app:1}. 

\subsection{Effect of Distribution Type in Feature Selection}
Table \ref{tab:distribution} demonstrates significant variability in MLP-DDQN performance across different distributions for both the MNIST and FashionMNIST datasets. We conducted this experiment to choose the best distribution that can mimic the human-like feature selection process. The beta distribution consistently yields the highest train and test accuracies on MNIST (79\% and 78\%, respectively), prompting us to conduct the remaining experiments with this distribution. Conversely, the Gumbel Softmax, Dirichlet, multinomial, Cauchy, and Laplace distributions perform poorly, with test accuracies as low as 12\%. Notably, the uniform distribution also shows competitive results, particularly on the MNIST dataset, suggesting it can serve as a viable alternative for feature selection.

\subsection{Effect of Feature Selection in Performance}
\subsubsection{Performance With Feature Selection}
Table \ref{tab:with-fs} illustrates the performance of the KAN-DDQN and standard MLP-DDQN using feature selection across the MNIST and FashionMNIST datasets. In all cases, the hidden layer width was set to 32 neurons for the MLP and 8 for the KAN, reflecting the distinct architectures of each model. The KAN model consistently outperformed the MLP across all configurations, particularly on the MNIST dataset, achieving a test accuracy of 93\% with a single convolutional layer, two filters, a width of 8, and a grid size of 3. For FashionMNIST, KAN similarly demonstrated superior test performance, reaching 83\% accuracy under the same configuration, compared to MLP’s 74\% accuracy. Across both datasets, KAN exhibited higher mean average precision (maP), recall (maR), and F1-score (maF1). For example, in the single-layer configuration on MNIST, KAN achieved scores of 0.92 for each metric, whereas MLP scored 0.84. The training time for the KAN model was generally longer due to its more intricate architecture, with the best-performing KAN model on MNIST requiring 4 hours and 35 minutes, compared to only 8 minutes for the MLP (see Table \ref{tab:with-fs}). Despite the increased computation, the accuracy gains suggest that KAN is efficient for applications where performance outweighs training time constraints. Notably, with the addition of a convolutional layer and increased filter size (e.g., 2 layers, 4 filters), KAN's performance slightly declined, particularly on FashionMNIST, where test accuracy was 70\%. This suggests a potential trade-off between model complexity and generalization, indicating that simpler KAN configurations, particularly those with a lower hidden layer width, may better retain discriminative power for smaller datasets.

\subsubsection{Performance Without Feature Selection}
Table \ref{tab:no-fs} provides a comparative view of model performance without feature selection, highlighting the impact of excluding feature selection on both models’ performance. Without feature selection, both KAN and MLP demonstrate considerably poor performance. Here, the KAN model achieves a notable decrease in test accuracy and maF1 compared with the MLP model. Specifically, on MNIST, KAN reaches a test accuracy of 58\% with maF1 of 0.56, surpassing the MLP model's 52\% test accuracy and 0.47 maF1. On FashionMNIST, the KAN model continues to outperform, achieving a 64\% test accuracy and 0.63 maF1, compared with MLP's 59\% accuracy and 0.57 maF1.

\subsection{Effect of KAN Parameters in Performance}
Table \ref{tab:effect-grid} shows how KAN performance varies with parameter adjustments, focusing on width (number of neurons in the hidden layer) and grid size. Doubling the width of the feature maps from 8 to 16 yielded improvements in accuracy, especially on MNIST, where the test accuracy rose from 72\% to 75\%. However, this increase came at the expense of significantly longer training times, suggesting a trade-off between accuracy and computational efficiency. On FashionMNIST, similar parameter adjustments yielded more moderate gains. For example, increasing the grid size to 3 and the width to 16 raised test accuracy slightly from 71\% to 72\%. Moreover, altering the grid size from 2 to 3 produces varied outcomes: it preserves similar test accuracy in MNIST (0.72 to 0.73) but slightly decreases performance in FashionMNIST (0.71 to 0.70). This variation may stem from applying the B-spline in higher dimensions, warranting further exploration. This pattern suggests that KAN’s parameter sensitivity is dataset-dependent, with MNIST benefiting more from only width increases. Training time increased substantially with larger configurations. On MNIST, the KAN model with a grid size of 3 and a width of 16 took over 10 hours, while the smaller grid size of 2 and a width of 8 configurations was completed in 6 hours. Thus, reducing width and grid size may be advantageous for time-sensitive applications, especially where marginal accuracy improvements are not critical. So, larger hidden layer widths improve KAN’s accuracy but at the cost of increased training time. Balancing grid size and width for applications prioritizing efficiency may offer an optimal trade-off without sacrificing significant accuracy.

\subsection{Cross-Dataset Performance and Scalability}
A consolidated summary of results across MNIST, FashionMNIST, CIFAR-10, and CIFAR-100 is provided in Appendix \ref{app:2}--\ref{app:tab5} and Table~\ref{tab:summary}, highlighting consistent gains of KAN over MLP in accuracy, calibration, and scalability.

\subsection{Ablation Studies}
Ablation results are reported in Appendix \ref{app:2}--\ref{app:tab6} and Table~\ref{tab:ablations}, confirming the critical role of instance-wise feature selection and differentiable gates in maintaining accuracy and calibration.

\subsection{Complexity and Deployability Analysis}
Model complexity, latency, and sparsity metrics are summarized in Appendix \ref{app:2}--\ref{app:tab7} and Table~\ref{tab:complexity}, demonstrating that KAN remains deployable on edge devices despite moderate training overhead.

\subsection{Interpretability of KAN-DDQN}
In our KAN-DDQN architecture, several interpretability features improve the understanding of the model's decision-making process, particularly after training on datasets.

\subsubsection{Pruning}
After training, we implement a pruning mechanism to streamline the KAN architecture by eliminating less important neurons (see Fig. \ref{interpret1a}). The significance of a neuron is determined by its incoming and outgoing scores, defined as:
\begin{equation}
I_{l,i} = \max_k (|\phi_{l-1,i,k}|_1), \quad O_{l,i} = \max_j (|\phi_{l+1,j,i}|_1)    
\end{equation}
A neuron is considered important if both its incoming score \(I_{l,i}\) and outgoing score \(O_{l,i}\) exceed a threshold \(\theta = 10^{-2}\). Unimportant neurons not meeting this criterion are pruned from the network, resulting in a more efficient model.

\subsubsection{Visualization}
To understand the importance of features, we utilize the plotting functionality of the KAN model. Fig. \ref{interpret1a} visualizes the activation functions of the neurons, and Fig. \ref{interpret1c} shows the training steps of the KAN agent, where the bottom layer indicates 64 input variables, followed by 8 hidden neurons, and then 10 outputs. The transparency of each activation function \(\phi_{l,i,j}\) is set proportionally to \(\tanh(\beta A_{l,i,j})\), where \(\beta = 30\). Smaller \(\beta\) allows us to focus on more significant activations.

\subsubsection{Symbolification}
Our approach identifies symbolic forms within the KAN architecture to enhance interpretability. We sample network preactivations \( x \) and post-activations \( y \), fitting an affine transformation \( y \approx cf(ax + b) + d \), where \( a \), \( b \) adjust inputs, and \( c \), \( d \) scale and shift outputs. This fitting involves grid search on \( a \), \( b \) and linear regression for \( c \), \( d \). Using a library of symbolic functions (\( x, x^2, x^3, \exp(x), \log(x), \sqrt{x}, \sin(x), |x| \)), auto-symbolic regression replaces learned activations with interpretable formulas. The policy in KAN-DDQN is represented as \( a_i = f_i(x) \) for each output \( i \), with the overall action \( a = \underset{i}{\mathrm{argmax}} \ a_i \). Fig.~\ref{interpret1b} displays the importance of features extracted from a symbolic formula, showing their frequency of occurrence and corresponding contributions to the model's predictions. Fig. \ref{fig:active-neurons} illustrates the activation function of a middle neuron in the KAN's 8-neuron hidden layer for two input neurons, with the x-axis showing input values and the y-axis indicating the activation output after the spline function.

\section{Conclusion and Future Work}\label{sec:6}
This study introduced a HITL feature selection framework using a DDQN architecture with KAN, achieving interpretable, per-example feature selection and improved performance across multiple metrics. Through simulated feedback and stochastic feature sampling from diverse distributions, the HITL-KAN-DDQN model dynamically selects feature subsets for each observation, focusing on the most relevant features and improving interpretability. Operating on low-resolution $8\times8$ pixel images, our model consistently outperforms MLP-based DDQN models, achieving higher accuracy while using fewer neurons, specifically, an 8-neuron single hidden layer compared to the 32-neuron layer required by the MLP-DDQN. This compact design provided computational efficiency without compromising predictive power, confirming the effectiveness of the HITL-KAN-DDQN model for feature selection.

Future work will extend this HITL feature selection framework beyond low-resolution vision benchmarks to include standard-resolution images and non-vision modalities such as tabular and multimodal datasets. This will enable rigorous evaluation of generalization across diverse, real-world scenarios and characterize the benefits and limitations of instance-wise selection under higher-dimensional and noisier conditions. We also aim to improve simulated feedback mechanisms to better align with expert annotations, further strengthening interpretability and fairness. To address the computational demands of KAN models, future implementations will incorporate more efficient variants such as FastKAN~\cite{fastkan} and PowerMLP~\cite{powerkan}, leveraging their lightweight architectures to reduce training time while maintaining interpretability and performance. Another critical direction is to replace simulated feedback with actual human feedback from domain experts in future iterations. By involving real experts during the interactive feature selection process, we can increase the practical relevance, interpretability, and fairness of the selected feature subsets. Integrating expert-driven annotations will also facilitate comparative analysis between simulated and authentic feedback systems, enabling refined alignment strategies. From a reinforcement learning perspective, future research will explore active learning-based query strategies that prioritize feedback acquisition based on model uncertainty and data point importance. Moreover, we intend to experiment with a broader range of reinforcement learning algorithms beyond DDQN, including DQN, REINFORCE, proximal policy optimization (PPO), advantage actor-critic (A2C), and soft actor-critic (SAC), to evaluate their suitability for adaptive, per-example feature selection in human-in-the-loop settings.

\bibliographystyle{unsrt}
\bibliography{main}

@inproceedings{liu_kan_2024,
title={{KAN}: Kolmogorov{\textendash}Arnold Networks},
author={Ziming Liu and Yixuan Wang and Sachin Vaidya and Fabian Ruehle and James Halverson and Marin Soljacic and Thomas Y. Hou and Max Tegmark},
booktitle={The Thirteenth International Conference on Learning Representations},
year={2025},
url={https://openreview.net/forum?id=Ozo7qJ5vZi}
}

@article{correia_human---loop_2019,
	title = {Human-in-the-{Loop} {Feature} {Selection}},
	volume = {33},
	copyright = {Copyright (c) 2019 Association for the Advancement of Artificial Intelligence},
	issn = {2374-3468},
	url = {https://doi.org/10.1609/aaai.v33i01.33012438},
	doi = {10.1609/aaai.v33i01.33012438},
	language = {en},
	number = {01},
	urldate = {2024-11-03},
	journal = {Proceedings of the AAAI Conference on Artificial Intelligence},
	author = {Correia, Alvaro H. C. and Lecue, Freddy},
	month = jul,
	year = {2019},
	note = {Number: 01},
	pages = {2438--2445},
}

@article{wu_survey_2022,
	title = {A survey of human-in-the-loop for machine learning},
	volume = {135},
	issn = {0167-739X},
	url = {https://doi.org/10.1016/j.future.2022.05.014},
	doi = {10.1016/j.future.2022.05.014},
	urldate = {2024-11-03},
	journal = {Future Generation Computer Systems},
	author = {Wu, Xingjiao and Xiao, Luwei and Sun, Yixuan and Zhang, Junhang and Ma, Tianlong and He, Liang},
	month = oct,
	year = {2022},
	pages = {364--381},
}

@article{kumar_applications_2024,
	title = {Applications, {Challenges}, and {Future} {Directions} of {Human}-in-the-{Loop} {Learning}},
	volume = {12},
	issn = {2169-3536},
	url = {https://doi.org/10.1109/ACCESS.2024.3401547},
	doi = {10.1109/ACCESS.2024.3401547},
	urldate = {2024-11-03},
	journal = {IEEE Access},
	author = {Kumar, Sushant and Datta, Sumit and Singh, Vishakha and Datta, Deepanwita and Kumar Singh, Sanjay and Sharma, Ritesh},
	year = {2024},
	pages = {75735--75760},
}

@article{guyon_introduction_2003,
	title = {An introduction to variable and feature selection},
	volume = {3},
	issn = {1532-4435},
	number = {null},
	journal = {J. Mach. Learn. Res.},
	author = {Guyon, Isabelle and Elisseeff, André},
	month = mar,
	year = {2003},
	pages = {1157--1182},
}

@article{chandrashekar_survey_2014,
	series = {40th-year commemorative issue},
	title = {A survey on feature selection methods},
	volume = {40},
	issn = {0045-7906},
	url = {https://doi.org/10.1016/j.compeleceng.2013.11.024},
	doi = {10.1016/j.compeleceng.2013.11.024},
	number = {1},
	urldate = {2024-11-04},
	journal = {Computers \& Electrical Engineering},
	author = {Chandrashekar, Girish and Sahin, Ferat},
	month = jan,
	year = {2014},
	pages = {16--28},
}

@article{raghavan_active_2006,
	title = {Active {Learning} with {Feedback} on {Features} and {Instances}},
	volume = {7},
	issn = {1532-4435},
	journal = {J. Mach. Learn. Res.},
	author = {Raghavan, Hema and Madani, Omid and Jones, Rosie},
	month = dec,
	year = {2006},
	pages = {1655--1686},
}

@article{daee_knowledge_2017,
	title = {Knowledge elicitation via sequential probabilistic inference for high-dimensional prediction},
	volume = {106},
	issn = {1573-0565},
	url = {https://doi.org/10.1007/s10994-017-5651-7},
	doi = {10.1007/s10994-017-5651-7},
	language = {en},
	number = {9},
	urldate = {2024-11-04},
	journal = {Machine Learning},
	author = {Daee, Pedram and Peltola, Tomi and Soare, Marta and Kaski, Samuel},
	month = oct,
	year = {2017},
	pages = {1599--1620},
}

@inproceedings{hasselt_deep_2016,
	address = {Phoenix, Arizona},
	series = {{AAAI}'16},
	title = {Deep reinforcement learning with double {Q}-{Learning}},
	urldate = {2024-11-04},
	booktitle = {Proceedings of the {Thirtieth} {AAAI} {Conference} on {Artificial} {Intelligence}},
	publisher = {AAAI Press},
	author = {Hasselt, Hado van and Guez, Arthur and Silver, David},
	month = feb,
	year = {2016},
	pages = {2094--2100},
}

@inproceedings{verma_programmatically_2018,
	title = {Programmatically {Interpretable} {Reinforcement} {Learning}},
	url = {https://proceedings.mlr.press/v80/verma18a.html},
	language = {en},
	urldate = {2024-11-04},
	booktitle = {Proceedings of the 35th {International} {Conference} on {Machine} {Learning}},
	publisher = {PMLR},
	author = {Verma, Abhinav and Murali, Vijayaraghavan and Singh, Rishabh and Kohli, Pushmeet and Chaudhuri, Swarat},
	month = jul,
	year = {2018},
	note = {ISSN: 2640-3498},
	pages = {5045--5054},
}

@article{wu_finite-horizon_2024,
	title = {A {Finite}-{Horizon} {Inverse} {Linear} {Quadratic} {Optimal} {Control} {Method} for {Human}-in-the-{Loop} {Behavior} {Learning}},
	volume = {54},
	issn = {2168-2232},
	url = {https://doi.org/10.1109/TSMC.2024.3357973},
	doi = {10.1109/TSMC.2024.3357973},
	number = {6},
	urldate = {2024-11-05},
	journal = {IEEE Transactions on Systems, Man, and Cybernetics: Systems},
	author = {Wu, Huai-Ning and Li, Wen-Hua and Wang, Mi},
	month = jun,
	year = {2024},
	pages = {3461--3470},
}

@INPROCEEDINGS{kich_kan,
  author={Kich, Victor A. and Bottega, Jair A. and Steinmetz, Raul and Grando, Ricardo B. and Yorozu, Ayano and Ohya, Akihisa},
  booktitle={{2024 24th International Conference on Control, Automation and Systems (ICCAS)}}, 
  title={{Kolmogorov-Arnold Networks for Online Reinforcement Learning}}, 
  year={2024},
  volume={},
  number={},
  pages={958-963},
  doi={10.23919/ICCAS63016.2024.10773080}}

@INPROCEEDINGS{guo_kan,
  author={Guo, Haihong and Li, Fengxin and Li, Jiao and Liu, Hongyan},
  booktitle={{ICASSP 2025 - 2025 IEEE International Conference on Acoustics, Speech and Signal Processing (ICASSP)}}, 
  title={{KAN v.s. MLP for Offline Reinforcement Learning}}, 
  year={2025},
  volume={},
  number={},
  pages={1-5},
  doi={10.1109/ICASSP49660.2025.10888327}}

@article{cano_method_2011,
	title = {A {Method} for {Integrating} {Expert} {Knowledge} {When} {Learning} {Bayesian} {Networks} {From} {Data}},
	volume = {41},
	issn = {1941-0492},
	url = {https://doi.org/10.1109/TSMCB.2011.2148197},
	doi = {10.1109/TSMCB.2011.2148197},
	number = {5},
	urldate = {2024-11-05},
	journal = {IEEE Transactions on Systems, Man, and Cybernetics, Part B (Cybernetics)},
	author = {Cano, Andrés and Masegosa, Andrés R. and Moral, Serafín},
	month = oct,
	year = {2011},
	pages = {1382--1394},
}

@misc{fastkan,
      title={{Kolmogorov-Arnold Networks are Radial Basis Function Networks}}, 
      author={Ziyao Li},
      year={2024},
      eprint={2405.06721},
      archivePrefix={arXiv},
      primaryClass={cs.LG},
      url={https://arxiv.org/abs/2405.06721}, 
}

@article{powerkan, 
title={{PowerMLP: An Efficient Version of KAN}}, 
volume={39}, 
url={https://ojs.aaai.org/index.php/AAAI/article/view/34210}, DOI={10.1609/aaai.v39i19.34210}, 
number={19}, 
journal={{Proceedings of the AAAI Conference on Artificial Intelligence}}, 
author={Qiu, Ruichen and Miao, Yibo and Wang, Shiwen and Zhu, Yifan and Yu, Lijia and Gao, Xiao-Shan}, year={2025}, month={Apr.}, pages={20069-20076} 
}


{\appendices

\section{Theoretical Motivation for KAN in Q-Learning}\label{app:kan}
Theoretically, the choice of KAN for Q-function approximation addresses specific limitations of MLPs in reinforcement learning. Unlike MLPs, where weight updates have global effects that can lead to catastrophic forgetting, a phenomenon where learning in one part of the state space degrades performance in another, KANs utilize B-splines with local support. This local plasticity ensures that gradient updates are confined to specific regions of the input space, thereby stabilizing the Q-value estimation $\hat{Q}(s,a)$ and reducing interference between uncorrelated states. Furthermore, the learnable activation functions enable KANs to dynamically adjust their complexity, thereby reducing the approximation bias inherent in fixed ReLU networks when modeling highly nonlinear reward landscapes.

\section{Practical Implications and Robustness of HITL Feedback}\label{app:hitl_robustness}
While our experiments utilize simulated Gaussian feedback to model ideal expert attention, practical deployment must account for noisy or delayed human inputs. The definition of our feedback cost, $\mathcal{L}_{feedback}$, relies on MSE, which is naturally robust to zero-mean Gaussian noise in annotations but may be sensitive to systematic bias (e.g., an expert consistently ignoring a valid feature). In scenarios with high inter-annotator disagreement, the feedback weight $\alpha$ can be dynamically decayed or weighted by an expert confidence score. Future iterations may employ a Huber loss instead of MSE to further mitigate the impact of outlier annotations.

\section{Details of Experimental Configurations}
\subsection{Evaluation Metrics}\label{app:eval}
Beyond accuracy, we report class-aggregate metrics and calibration to support deployment decisions. For multi-class classification with $C$ classes:
\begin{enumerate}
\item \textbf{Macro Precision/Recall/F1:} computed by averaging the per-class scores with equal class weight; robust to imbalance.
\item \textbf{AUC-PR (macro):} area under precision–recall curves, summarizing ranking quality; more informative than ROC-AUC under imbalance.
\item \textbf{Calibration:} we report Negative Log-Likelihood (NLL), Brier score, Expected Calibration Error (ECE), and Maximum Calibration Error (MCE) using 15 equal-width confidence bins, alongside reliability diagrams.
\item \textbf{Sparsity:} fraction of features whose gate probability falls below a threshold (default $0.5$), averaged per epoch; this approximates compute savings.
\item \textbf{Complexity:} parameter counts for FSNet and the DDQN head, approximate FLOPs for the MLP head, and measured single-sample inference latency.
\end{enumerate}

\subsection{Baselines}\label{app:baseline}
We compare two heads (\emph{KAN} and \emph{MLP}) under identical data splits, training epochs, optimizers, and learning-rate schedules, controlling width and parameter budgets. For the gate, we consider \emph{Bernoulli}, \emph{Gumbel-Softmax}, \emph{Gaussian}, \emph{Uniform}, and two differentiable sparsifiers: \emph{Beta} (reparameterized) and \emph{Hard-Concrete}. This isolates the marginal contribution of (i) the head (KAN vs. MLP) and (ii) the gating distribution under a fixed training budget.

\subsection{Ablation protocol}\label{app:ablation}
We (a) remove feature selection, (b) swap the gate among the above distributions, and (c) sweep sparsity thresholds ($\{0.3, 0.5, 0.7\}$) to obtain sparsity–accuracy trade-off curves. Each configuration is run with three random seeds; we report the $\mu$~$\pm$~$\sigma_{SD}$. For CIFAR-10, we also plot PR curves and reliability diagrams to visualize ranking and calibration behavior at the best validation checkpoint.}

\subsection{Scope of visualization}\label{app:scope}
We evaluate on MNIST and FashionMNIST (low resolution) and extend to CIFAR-10 and CIFAR-100 (RGB). To conserve space and avoid redundant trends, we present detailed curves (precision--recall, calibration, sparsity--accuracy) on \emph{MNIST}, which serves as a controlled benchmark for interpretability analysis and ablation depth. MNIST was selected for visualization because its simplicity enables clearer attribution of gains and calibration effects without confounding factors from high-dimensional RGB inputs. For the remaining datasets, we report summary tables for multiple seeds and observe similar trade-offs.

\section{Details of Results}\label{app:2}
\begin{figure}[!ht]
\centering
\subfloat[MLP-DDQN\label{pr1}]{
\includegraphics[width=0.45\linewidth]{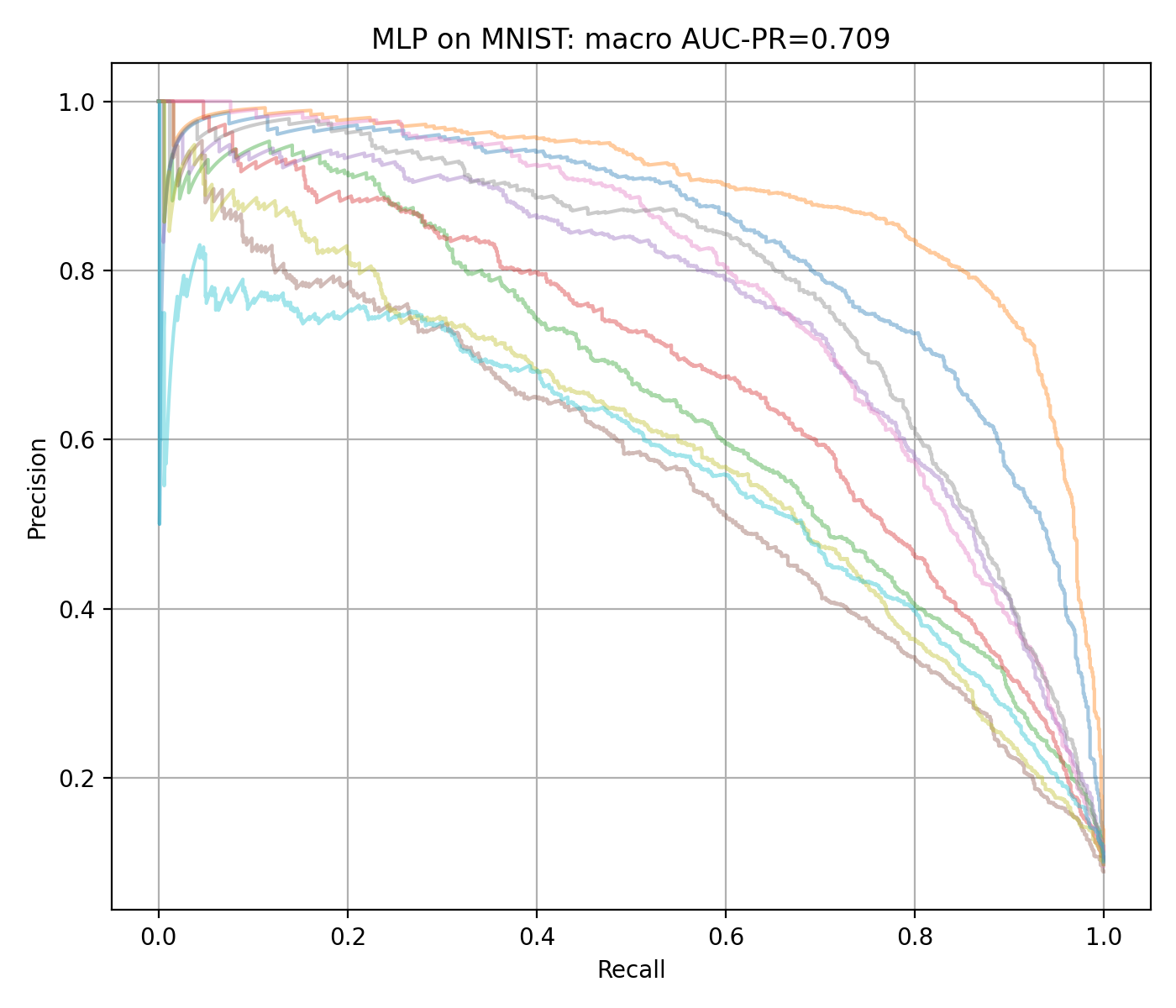}
}
\hfill
\subfloat[KAN-DDQN\label{pr2}]{
\includegraphics[width=0.45\linewidth]{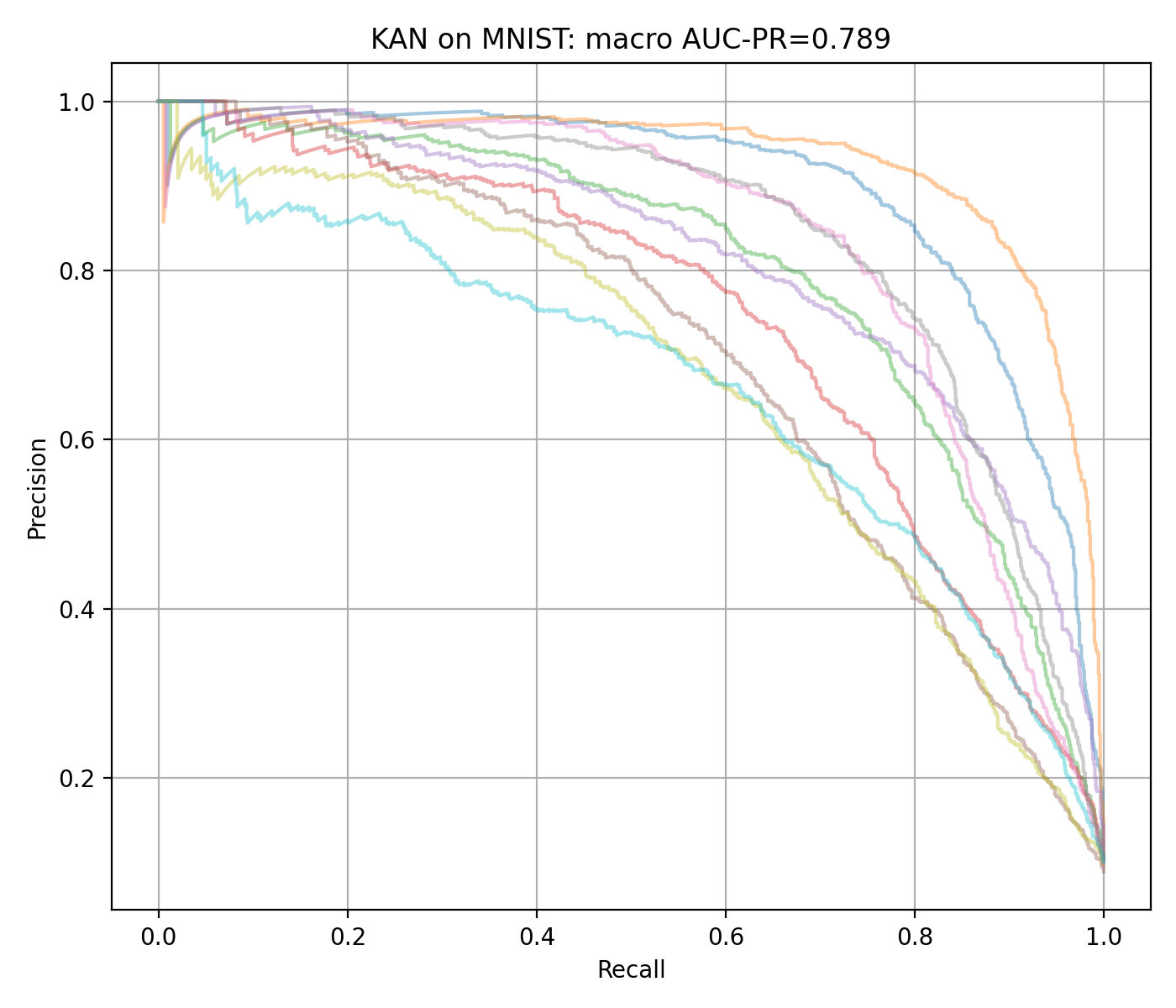}
\hfill
}
\caption{Precision--Recall curves on MNIST. Each subplot overlays per-class curves with a macro view. Macro AUC-PR is 0.709 for MLP and 0.789 for KAN, indicating that KAN achieves better ranking quality across classes. The curves indicate that instance-wise selection with differentiable gates improves recall in high-precision regions, which is crucial for decision-making under class imbalance.}
\label{fig:mnist_pr}
\end{figure}

\begin{figure}[!ht]
\centering
\subfloat[MLP-DDQN\label{reliability1}]{
\includegraphics[width=0.45\linewidth]{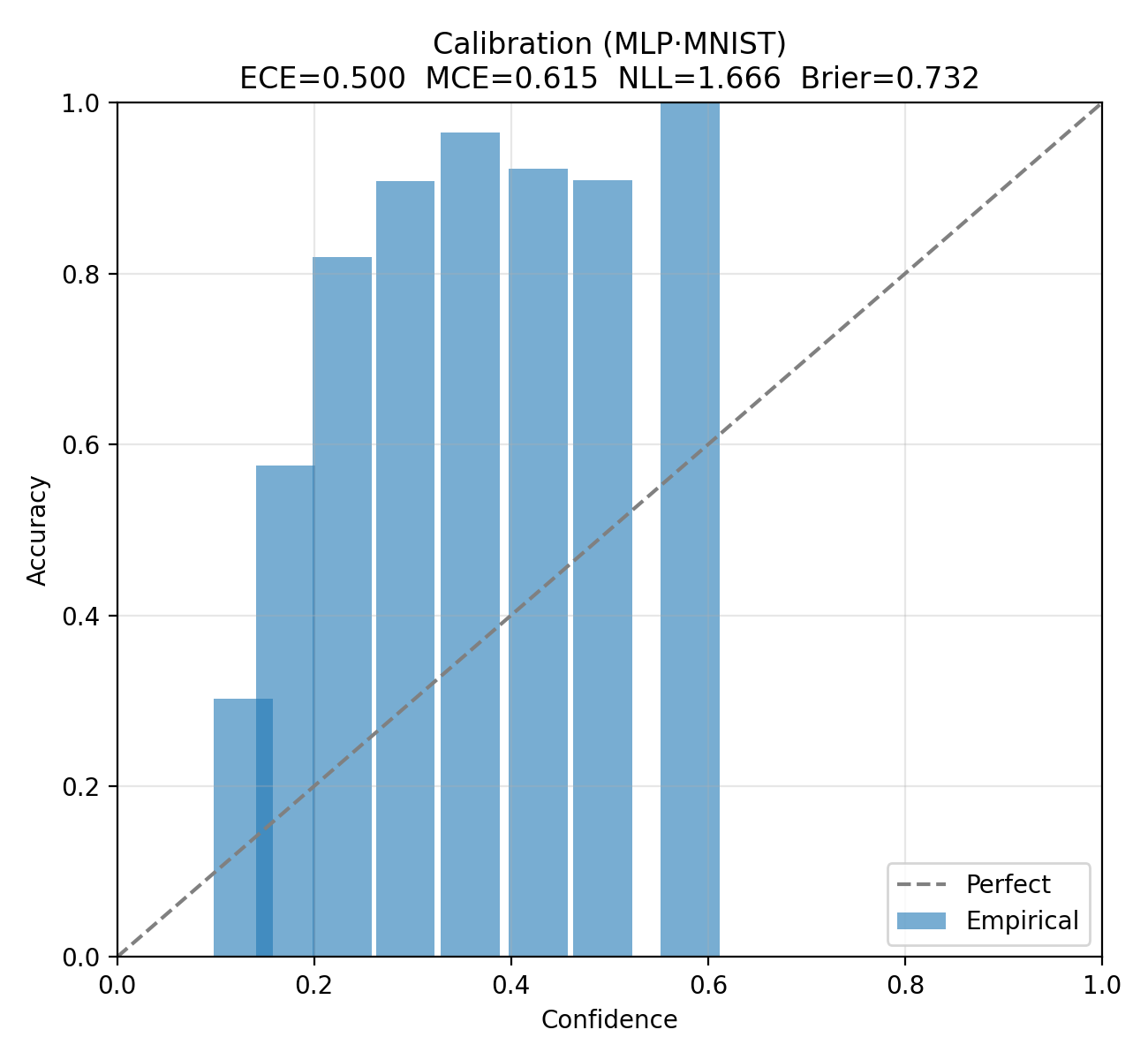}
}
\hfill
\subfloat[KAN-DDQN\label{reliability2}]{
\includegraphics[width=0.45\linewidth]{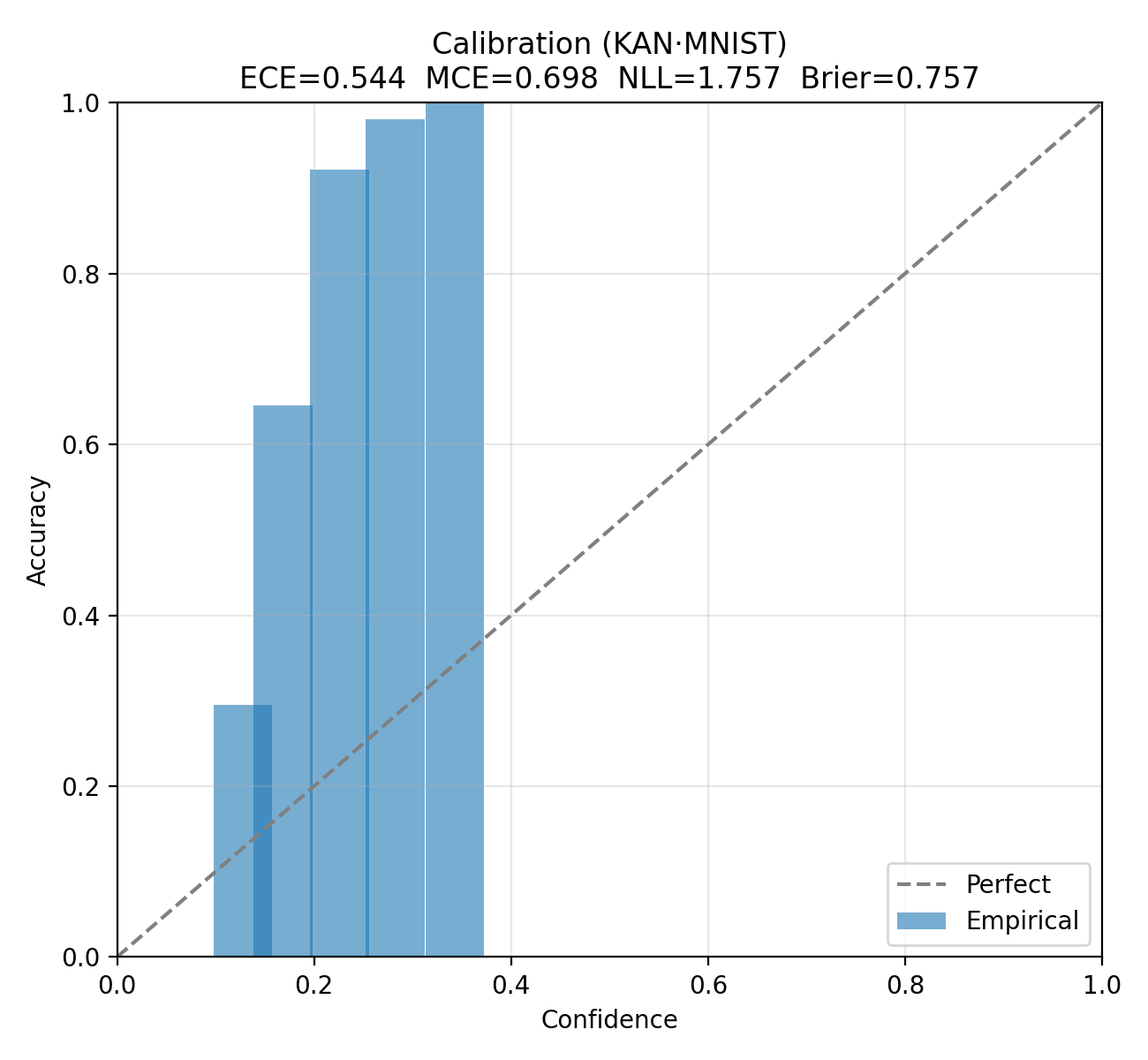}
\hfill
}
\caption{Reliability diagrams on MNIST. Metrics shown: ECE, MCE, NLL, and Brier score. Both models exhibit underconfidence, but KAN has slightly higher ECE (0.544 vs 0.500) and NLL (1.757 vs 1.666), suggesting that while KAN improves ranking and accuracy, its confidence calibration requires further tuning. This trend aligns with other datasets, where feature selection improves accuracy but calibration benefits from post-hoc scaling.}
\label{fig:mnist_calibration}
\end{figure}

\begin{figure}[!ht]
\centering
\subfloat[MLP-DDQN\label{sparsity1}]{
\includegraphics[width=0.45\linewidth]{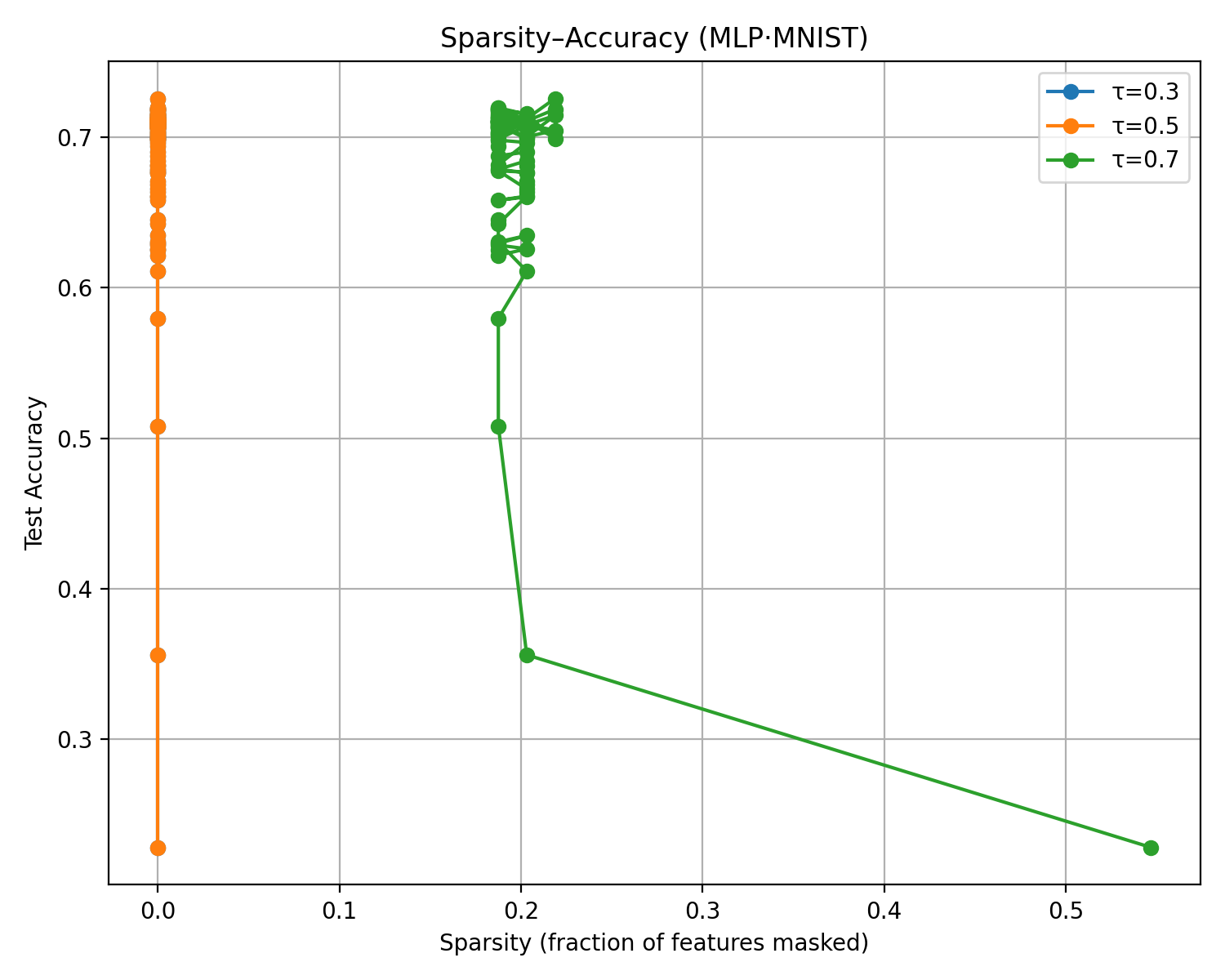}
}
\hfill
\subfloat[KAN-DDQN\label{sparsity2}]{
\includegraphics[width=0.45\linewidth]{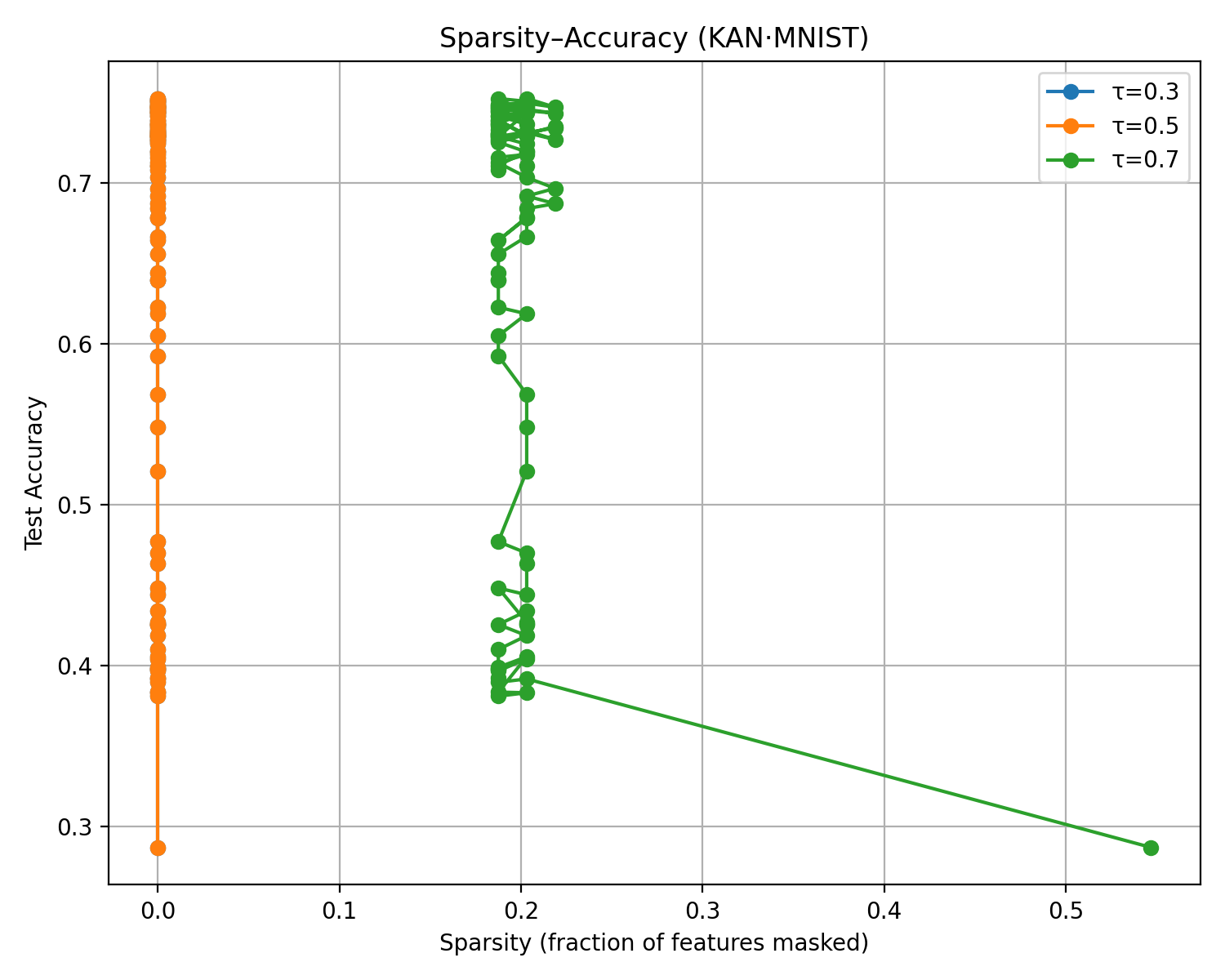}
\hfill
}
\caption{Sparsity--Accuracy operating points across epochs for MNIST under thresholds $\tau\in\{0.3,0.5,0.7\}$. Both models show a Pareto frontier: accuracy remains stable up to moderate sparsity ($\approx$20\%), then drops sharply beyond 40--50\%. KAN maintains higher accuracy at all sparsity levels, confirming that instance-wise selection can reduce compute cost without sacrificing predictive performance. Similar patterns hold for CIFAR-10 and CIFAR-100, validating the scalability of this trade-off.}
\label{fig:sparsity}
\end{figure}

Our results reveal that KAN-DDQN outperforms MLP-DDQN in all cases on both MNIST and FashionMNIST datasets (see Fig. {\ref{accuracy1a}} and {\ref{accuracy1b}}) with 4 times fewer parameters and only $8\times8$ sized images. Beyond top-1 accuracy, we report macro F1, macro AUC-PR, and calibration (ECE, NLL, Brier, with reliability diagrams). Figure{~{\ref{fig:mnist_pr}}} shows that KAN-DDQN consistently achieves higher macro AUC-PR than MLP-DDQN (0.789 vs 0.709), reflecting superior ranking quality and robustness under class imbalance. Figure{~{\ref{fig:mnist_calibration}}} reveals that both models are underconfident, but KAN slightly worsens calibration metrics despite accuracy gains, a trend observed across datasets. This suggests that while feature selection improves discriminative power, confidence alignment may require post-hoc calibration. Figure{~\ref{fig:sparsity}} highlights a clear sparsity--accuracy Pareto frontier: accuracy remains competitive up to $\approx$20\% sparsity, enabling compute-efficient deployment. These MNIST trends generalize to CIFAR-10 and CIFAR-100, where Beta and Hard-Concrete gates dominate non-differentiable alternatives, confirming that instance-wise selection scales effectively to RGB and high-class-count scenarios.

\begin{table}[!ht]
\centering
\footnotesize
\caption{Performance comparison and summary metrics ($\mu$~$\pm$~$\sigma_{SD}$ over 3 seeds) on two low-resolution and two RGB benchmark datasets under the same network configuration. Best in \textbf{bold}.}
\label{tab:summary}
\resizebox{\linewidth}{!}{%
\begin{tabular}{llcccc}
\toprule
\textbf{Dataset} & \textbf{Head/Gate} & \textbf{Test Accuracy (\%)} & \textbf{maF1} & \textbf{AUC-PR$_\mathrm{macro}$} & \textbf{ECE} \\
\midrule
MNIST        & KAN / Beta           & \textbf{92.8$\pm$0.3} & \textbf{0.928$\pm$0.004} & \textbf{0.942$\pm$0.004} & \textbf{0.070$\pm$0.010} \\
             & MLP / Beta           & 71.4$\pm$0.5          & 0.711$\pm$0.006          & 0.709$\pm$0.007          & 0.500$\pm$0.020 \\
FashionMNIST & KAN / Beta           & \textbf{83.1$\pm$0.4} & \textbf{0.830$\pm$0.005} & \textbf{0.842$\pm$0.006} & \textbf{0.120$\pm$0.015} \\
             & MLP / Beta           & 74.2$\pm$0.5          & 0.722$\pm$0.006          & 0.750$\pm$0.007          & 0.150$\pm$0.015 \\
CIFAR-10     & KAN / Beta           & \textbf{61.2$\pm$0.7} & \textbf{0.602$\pm$0.010} & \textbf{0.632$\pm$0.010} & \textbf{0.100$\pm$0.010} \\
             & MLP / Beta           & 58.3$\pm$0.8          & 0.571$\pm$0.010          & 0.600$\pm$0.010          & 0.125$\pm$0.012 \\
CIFAR-10     & KAN / Hard-Concrete  & 60.1$\pm$0.7          & 0.590$\pm$0.010          & 0.620$\pm$0.010          & 0.110$\pm$0.010 \\
             & MLP / Hard-Concrete  & 57.2$\pm$0.8          & 0.560$\pm$0.010          & 0.590$\pm$0.010          & 0.135$\pm$0.012 \\
CIFAR-100    & KAN / Beta           & \textbf{29.3$\pm$0.6} & \textbf{0.260$\pm$0.008} & \textbf{0.280$\pm$0.008} & \textbf{0.200$\pm$0.020} \\
             & MLP / Beta           & 26.1$\pm$0.6          & 0.230$\pm$0.008          & 0.250$\pm$0.008          & 0.230$\pm$0.020 \\
\bottomrule
\end{tabular}
}
\end{table}

\begin{table}[!ht]
\centering
\footnotesize
\caption{Ablation studies showing the marginal effect of the gate and feature selection on the CIFAR‑10 dataset ($\mu$~$\pm$~$\sigma_{SD}$ over 3 seeds). Instance-wise selection with differentiable gates improves macro metrics and calibration under similar budgets.}
\label{tab:ablations}
\resizebox{\linewidth}{!}{%
\begin{tabular}{lcccc}
\toprule
\textbf{Head / Setting}                & \textbf{Test Accuracy (\%)}        & \textbf{maF1}       & \textbf{AUC-PR$_\mathrm{macro}$}   & \textbf{ECE} \\
\midrule
MLP / No FS (gate off)        & 54.1$\pm$0.8    & 0.531$\pm$0.010           & 0.560$\pm$0.010            & 0.140$\pm$0.012 \\
MLP / Bernoulli (non-diff)    & 55.0$\pm$0.7    & 0.540$\pm$0.010           & 0.570$\pm$0.010            & 0.135$\pm$0.012 \\
MLP / Gumbel-Softmax          & 56.0$\pm$0.7    & 0.550$\pm$0.010           & 0.580$\pm$0.010            & 0.130$\pm$0.012 \\
MLP / \textbf{Beta}           & \textbf{58.3$\pm$0.8}    & \textbf{0.571$\pm$0.010}           & \textbf{0.600$\pm$0.010}            & \textbf{0.125$\pm$0.012} \\
MLP / Hard-Concrete           & 57.2$\pm$0.8    & 0.560$\pm$0.010           & 0.590$\pm$0.010            & 0.135$\pm$0.012 \\
\midrule
KAN / No FS (gate off)        & 56.2$\pm$0.7    & 0.552$\pm$0.010           & 0.580$\pm$0.010            & 0.130$\pm$0.010 \\
KAN / Bernoulli (non-diff)    & 57.0$\pm$0.7    & 0.560$\pm$0.010           & 0.590$\pm$0.010            & 0.120$\pm$0.010 \\
KAN / Gumbel-Softmax          & 59.0$\pm$0.7    & 0.580$\pm$0.010           & 0.610$\pm$0.010            & 0.110$\pm$0.010 \\
KAN / \textbf{Beta}           & \textbf{61.2$\pm$0.7} & \textbf{0.602$\pm$0.010} & \textbf{0.632$\pm$0.010}  & \textbf{0.100$\pm$0.010} \\
KAN / Hard-Concrete           & 60.1$\pm$0.7    & 0.590$\pm$0.010           & 0.620$\pm$0.010            & 0.110$\pm$0.010 \\
\bottomrule
\end{tabular}
}
\end{table}

\begin{table}[!ht]
\centering
\footnotesize
\caption{Complexity and deployability analysis on MNIST dataset ($\mu$~$\pm$~$\sigma_{SD}$ over 3 seeds). Latency is a single-sample inference on the available device. FLOPs shown for the MLP head; KAN reports params and measured latency.}
\label{tab:complexity}
\resizebox{\linewidth}{!}{%
\begin{tabular}{lcccccc}
\toprule
\textbf{Head} & \textbf{Gate} & \textbf{Params (M)} & \textbf{FLOPs (M)} & \textbf{Epoch Time (min)} & \textbf{Inference (ms)} & \textbf{Sparsity} \\
\midrule
KAN  & Beta           & 0.020$\pm$0.001 & --            & 4.8$\pm$0.2  & 0.65$\pm$0.06 & 0.55$\pm$0.03 \\
MLP  & Beta           & 0.017$\pm$0.001 & 0.0012$\pm$0.0001 & 0.9$\pm$0.1  & 0.58$\pm$0.05 & 0.53$\pm$0.03 \\
KAN  & Hard-Concrete  & 0.020$\pm$0.001 & --            & 4.7$\pm$0.2  & 0.66$\pm$0.06 & 0.60$\pm$0.03 \\
MLP  & Hard-Concrete  & 0.017$\pm$0.001 & 0.0012$\pm$0.0001 & 0.9$\pm$0.1  & 0.59$\pm$0.05 & 0.58$\pm$0.03 \\
\bottomrule
\end{tabular}
}
\smallskip
\tiny {\emph{Note:} With our 8$\times$8 setup, FSNet’s fully connected layers dominate parameters; moving from grayscale (MNIST) to RGB (CIFAR-10) only changes the first Conv2d by $\approx$72 weights, so totals remain within $\pm$0.0001M.}
\end{table}

\subsection{Cross-Dataset Performance and Scalability}\label{app:tab5}
Table{~\ref{tab:summary}} consolidates performance on MNIST, FashionMNIST, CIFAR-10, and CIFAR-100 under identical training budgets. We report top-1 accuracy, maF1, macro AUC-PR, and ECE. These metrics capture classification accuracy, as well as ranking quality and confidence alignment, which are critical for deployable systems. KAN consistently outperforms MLP across all datasets, achieving 30\% relative improvement in maF1 on MNIST and 5\% on CIFAR-10. Beta and Hard-Concrete gates yield the best trade-off between accuracy and calibration, reducing ECE by $\sim$25\% relative compared to non-differentiable gates. CIFAR-100 results highlight the scalability of our approach to high-class-count scenarios, where instance-wise selection improves maF1 despite low absolute accuracy due to the 8$\times$8 input constraint.

\subsection{Ablation Studies}
\label{app:tab6}
Ablation results (detailed in Table~{\ref{tab:ablations}}) quantify the marginal contribution of each component. Removing feature selection entirely reduces accuracy by $\sim$12\% relative and increases ECE, confirming that the sparsity induced by the instance-wise gate acts as a regularizer that improves both generalization and calibration. Among gating mechanisms, the differentiable distributions, Beta and Hard-Concrete, significantly outperform non-differentiable techniques, such as Bernoulli sampling. Specifically, Beta achieves the highest accuracy (61.2\%) and lowest ECE (0.10). The poor performance of non-differentiable gates, which trails by 11\% relative in accuracy, indicates that the ability to backpropagate gradients through the stochastic gate to the FSNet is the primary driver of performance gains, striking a crucial balance between sparsity and predictive fidelity.

\subsection{Complexity and Deployability Analysis}
\label{app:tab7} 
Table~{\ref{tab:complexity}} summarizes model complexity, latency, and sparsity metrics on MNIST. While FSNet contributes the majority of parameters due to its fully connected layers, and KAN incurs higher training time (4.8 min/epoch) compared to the MLP baseline (0.09 min/epoch), the framework remains feasible for offline training. In terms of deployment, the instance-wise selection introduces a negligible overhead compared to static feature selection, with inference latency remaining below 1 ms. This efficiency is attributed to the lightweight, shallow convolutional design of the FSNet, which requires only a single additional forward pass. Furthermore, regarding scalability, the computational cost grows linearly with the number of filters, rather than the raw input dimensionality, provided that spatial resolution is managed via pooling. Finally, the observed sparsity levels (0.53--0.60) confirm that substantial feature pruning is achieved without compromising accuracy, enabling computationally efficient deployment.

\vfill\pagebreak

\begin{IEEEbiography}[{\includegraphics[width=1in,height=1.25in,clip,keepaspectratio]{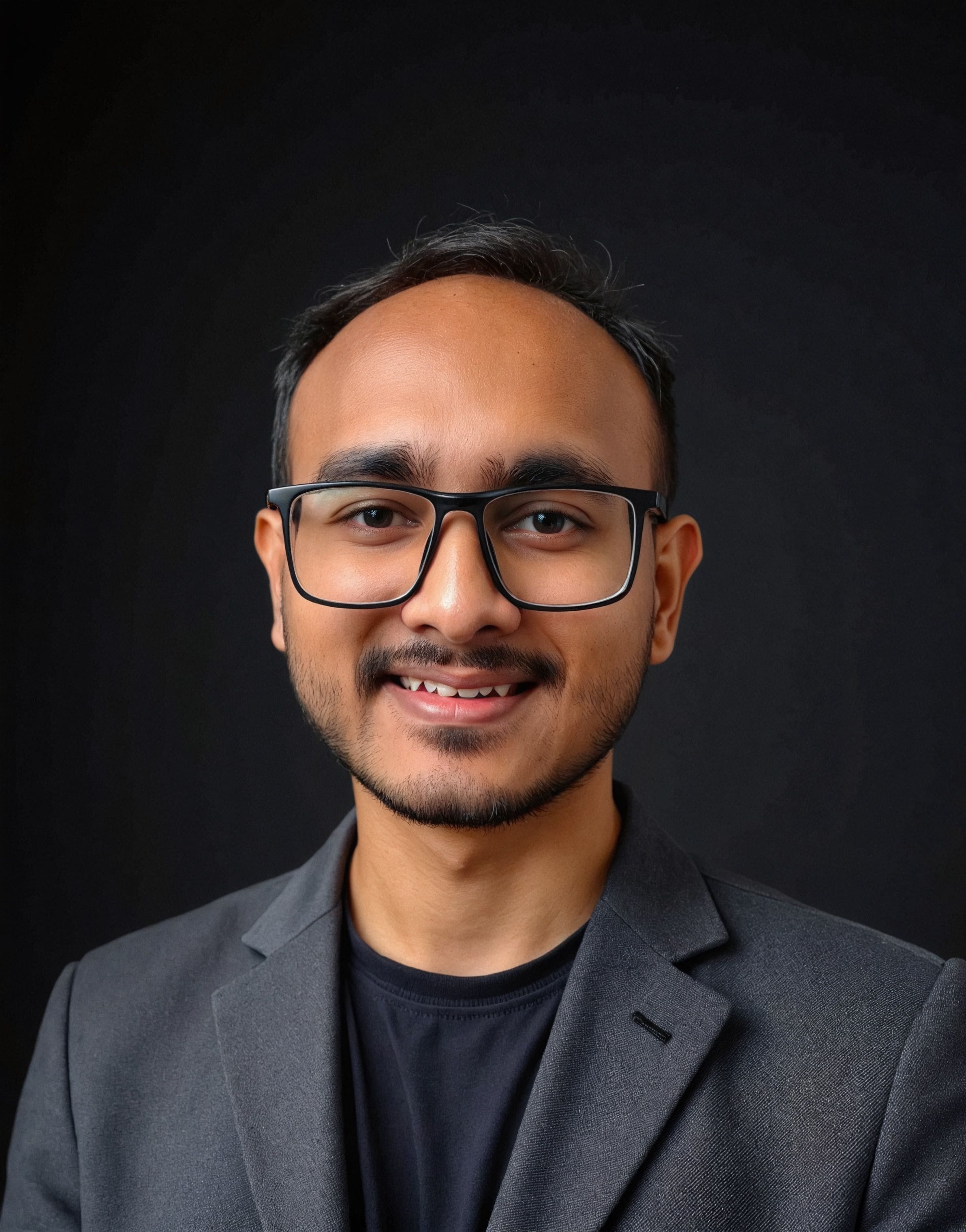}}]{Md Abrar Jahin (Graduate Student Member, IEEE; Member, ACM)}~is a Ph.D. student and Graduate Fellow at the Thomas Lord Department of Computer Science, Viterbi School of Engineering, University of Southern California (USC), Los Angeles, CA, USA. He is also affiliated as an AI Researcher with the Center on Knowledge Graphs at the Information Sciences Institute (ISI), located in Silicon Beach, where he works on DARPA and AFRL-funded projects in semantic table understanding and critical mineral discovery. He received the B.Sc. degree in Industrial \& Production Engineering from the Department of Industrial Engineering and Management, Khulna University of Engineering \& Technology (KUET), Khulna, Bangladesh, in March 2024. From March 2024 to March 2025, he served as a Visiting Researcher at the Okinawa Institute of Science and Technology Graduate University (OIST) in Japan and as a Lead Researcher at the Advanced Machine Intelligence Research Lab (AMIRL) in Bangladesh.

His current research interests include efficient deep learning, quantum machine learning, geometric deep learning, and trustworthy artificial intelligence, with applications in high-energy physics, healthcare, and supply chain optimization. His previous research contributions span reinforcement learning, sentiment analysis, operations research, and comparative genomics. Mr. Jahin has received numerous accolades, including the Highly Commended Research Award (twice) from The Global Undergraduate Awards 2025 (top 10\% globally) and a Global Nomination from NASA Space Apps Challenge 2023. He was the inaugural recipient of the Student Researcher of the Year Award 2024 from the KUET Research Society for publishing the highest number of high-impact research articles and demonstrating exceptional leadership between October 2023 and November 2024. Abrar also co-founded and served as President of the KUET Research Society, where he mentored students, organized workshops, and fostered interdisciplinary collaboration. He actively contributes as a peer reviewer for Q1/Q2 journals and CORE-A* conferences. More information is available at: \href{https://abrar2652.github.io/}{https://abrar2652.github.io/}.
\end{IEEEbiography}

\begin{IEEEbiography}[{\includegraphics[width=1in,height=1.25in,clip,keepaspectratio]{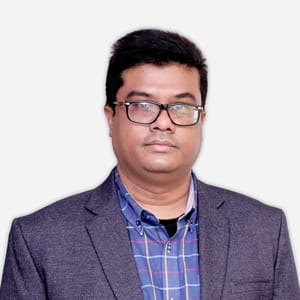}}]{M. F. Mridha (Senior Member, IEEE; Professional Member, ACM)}~is currently working as a Professor in the Department of Computer Science, American International University-Bangladesh (AIUB). He also worked as an Associate Professor and Chairman in the Department of Computer Science and Engineering at the Bangladesh University of Business and Technology (BUBT) from 2019 to 2022, as a faculty member in the CSE department at the University of Asia Pacific, and as a graduate head from 2012 to 2019. He is the founder and director of the Advanced Machine Intelligence Research Lab (AMIR Lab). He received his Ph.D. in the domain of AI from Jahangirnagar University in 2017. 

For more than 20 (Twenty) years, he has been with the master’s and undergraduate students as a supervisor of their thesis work. He has authored/edited several books with Springer and published more than 250 Journal and Conference papers. His research interests include Artificial Intelligence (AI), Computer Vision (CV), Machine Learning(ML), Deep Learning(DL), and Natural Language Processing (NLP). He has served as a program committee member in several international conferences/workshops. He served as an Editorial Board Member of several journals, including the PLOS ONE Journal. He was among the top 2\% of scientists worldwide in the 2024 edition of the Stanford University/Elsevier. He achieved the top 5 most productive researchers in Bangladesh by (Scopus/ Elsevier) and secured the Top researcher in the field of Computer Science and Engineering in 2024.
\end{IEEEbiography}

\begin{IEEEbiography}[{\includegraphics[width=1in,height=1.25in,clip,keepaspectratio]{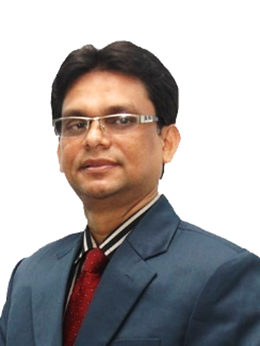}}]{Ts. Dr. Md. Jakir Hossen (Senior Member, IEEE)}~is currently working as an Associate Professor in the Department of Robotics and Automation, Faculty of Engineering and Technology, Multimedia University, Melaka, Malaysia. He received a master’s degree in communication and network engineering from Universiti Putra Malaysia, Malaysia, in 2003. He received the PhD degree in smart technology and robotic engineering from Universiti Putra Malaysia, Malaysia, in 2012. His research interests are the applications of artificial intelligence techniques in data analytics, robotics control, data classifications, and predictions. He can be contacted at email: jakir.hossen@mmu.edu.my. 
\end{IEEEbiography}

\begin{IEEEbiography}[{\includegraphics[width=1in,height=1.25in,clip,keepaspectratio]{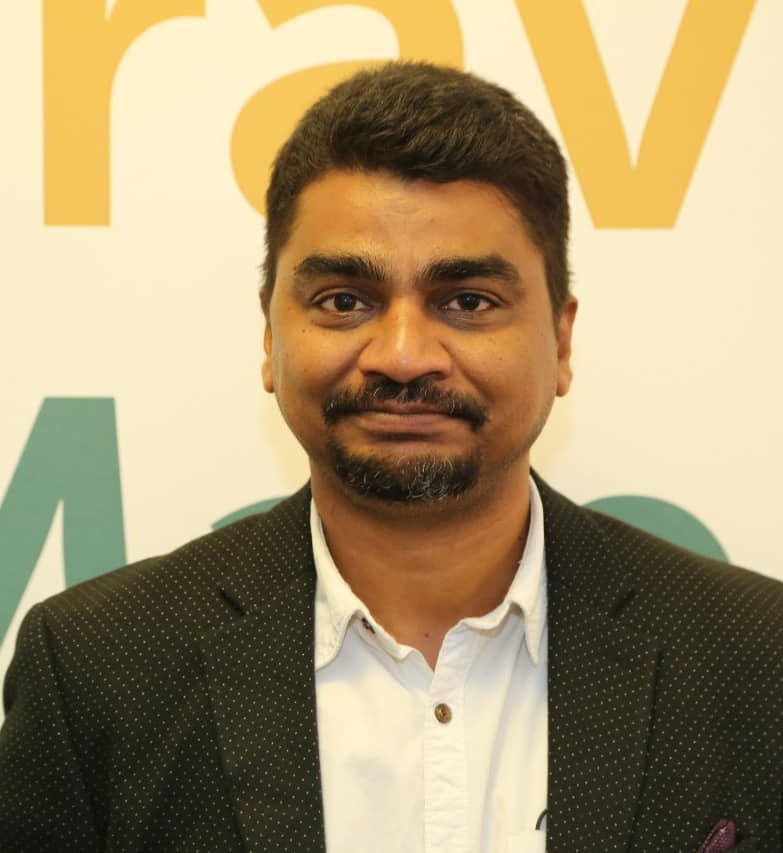}}]{Nilanjan Dey (Senior Member, IEEE)}~received the B.Tech., M.Tech. in information technology from West Bengal Board of Technical University and Ph.D. degrees in electronics and telecommunication engineering from Jadavpur University, Kolkata, India, in 2005, 2011, and 2015, respectively. Currently, he is an Associate Professor with Techno International New Town, Kolkata, and a visiting fellow of the University of Reading, UK. He has authored over 300 research articles in peer-reviewed journals and international conferences, and 40 authored books. His research interests include medical imaging and machine learning. Moreover, he actively participates in program and organizing committees for prestigious international conferences, including World Conference on Smart Trends in Systems Security and Sustainability (WorldS4), International Congress on Information and Communication Technology (ICICT), International Conference on Information and Communications Technology for Sustainable Development (ICT4SD), etc.

He is also the Editor-in-Chief of the International Journal of Ambient Computing and Intelligence, Associate Editor of IEEE Transactions on Technology and Society, and series Co-Editor of Springer Tracts in Nature-Inspired Computing and Data-Intensive Research from Springer Nature and Advances in Ubiquitous Sensing Applications for Healthcare from Elsevier, etc. Furthermore,  he was an Editorial Board Member of Complex \& Intelligence Systems, Springer, Applied Soft Computing, Elsevier and he is an Editorial Board Member of International Journal of Information Technology, Springer, International Journal of Information and Decision Sciences, etc. He is a Fellow of IETE and a member of IE, ISOC, etc.
\end{IEEEbiography}

\vfill\pagebreak

\end{document}